\documentclass[preprint]{elsarticle}

\usepackage{graphicx}
\usepackage{float}
\usepackage{amssymb}
\usepackage{amsmath}
\usepackage{soul}
\usepackage{lineno}

\usepackage{setspace,adjustbox}
\usepackage{verbatim}
\usepackage{subfigure}
\usepackage{bbm}
\usepackage{textcomp}
\usepackage{multirow}
\usepackage[margin=2.5cm]{geometry}
\usepackage[skip=0.5\baselineskip]{caption}
\usepackage[colorlinks]{hyperref}
\usepackage[table,dvipsnames]{xcolor}
\definecolor{myblue1}{rgb}{0.1, 0.3, 0.5}
\definecolor{myred1}{rgb}{0.8, 0, 0}
\definecolor{mypink3}{cmyk}{0, 0.7808, 0.4429, 0.1412}
\definecolor{mygray}{gray}{0.6}
\usepackage{array}
\newcolumntype{L}{>{\centering\arraybackslash}m{3cm}}
\newcolumntype{P}[1]{>{\centering\arraybackslash}p{#1}}
\newcolumntype{M}[1]{>{\centering\arraybackslash}m{#1}}
\AtBeginDocument{\hypersetup{colorlinks, urlcolor = myblue1, anchorcolor= myblue1, linkcolor = myred1, citecolor = myblue1}}

\usepackage[english]{babel}
\usepackage{graphicx, inputenc, stackengine}
\graphicspath{ {./figures/} }
\usepackage{chemfig,natbib,changepage}
\usepackage{framed}
\usepackage[normalem]{ulem}
\usepackage{amsmath}
\usepackage{amsthm}
\usepackage{tikz}
\usepackage{amssymb}
\usepackage{amsfonts}
\usepackage{enumerate}
\usepackage{tabu}
\usepackage[version=4]{mhchem}

\theoremstyle{definition}

\journal{Computers and Chemical Engineering}

\usepackage[most]{tcolorbox}
\newtcolorbox{highlighted}{colback=yellow,coltext=red,breakable}

\begin{document}

\begin{frontmatter}


% Following are the possible titles:
% \title{Language Models in Science \& Engineering -- Quo Vadis?}
\title{\textit{Quo Vadis} ChatGPT?\\ From Large Language Models to Large Knowledge Models}

% \title{Intelligent Mechanistic Explanation Generator -- IMEXGEN: \\Part 1. Automated Knowledge Organization}


% %% use optional labels to link authors explicitly to addresses:
% \author{Venkat Venkatasubramanian\corref{cor1}}
% \cortext[cor1]{Corresponding author}
% \ead{venkat@columbia.edu}

% \author{Arijit Chakraborty}
% \ead{ac4758@columbia.edu}

\author{\stackunder{Venkat Venkatasubramanian$^{*}$ and Arijit Chakraborty}}
\cortext[cor1]{Corresponding author}
  % {\texttt{\{au1, au2\}@email-adress.com}}}

\makeatletter \emailauthor{venkat@columbia.edu}{Venkat Venkatasubramanian} \emailauthor{ac4758@columbia.edu}{Arijit Chakraborty} \let\emailauthor\@gobbletwo \makeatother

\address{Complex Resilient Intelligent Systems Laboratory, Department of Chemical Engineering, Columbia University, New York, NY 10027, USA}

\begin{abstract}
The startling success of ChatGPT and other large language models (LLMs) using transformer-based generative neural network architecture in applications such as natural language processing and image synthesis has many researchers excited about potential opportunities in process systems engineering (PSE). The almost human-like performance of LLMs in these areas is indeed very impressive, surprising, and a major breakthrough. Their capabilities are very useful in certain tasks, such as writing first drafts of documents, code writing assistance, text summarization, etc. However, their success is limited in highly scientific domains as they cannot yet reason, plan, or explain due to their lack of in-depth domain knowledge. This is a problem in domains such as chemical engineering as they are governed by fundamental laws of physics and chemistry (and biology), constitutive relations, and highly technical knowledge about materials, processes, and systems. Although purely data-driven machine learning has its immediate uses, the long-term success of AI in scientific and engineering domains would depend on developing hybrid AI systems that use first principles and technical knowledge effectively. We call these hybrid AI systems \textit{Large Knowledge Models} (LKMs), as they will not be limited to only NLP-based techniques or NLP-like applications. In this paper, we discuss the challenges and opportunities in developing such systems in chemical engineering.  
\end{abstract}


\end{frontmatter}

\section{Introduction}\label{sec:Intro}

In recent years, the formidable combination of large data sets, new machine learning (ML) algorithms and architectures, and powerful hardware has led to the birth of large language models (LLMs), such as ChatGPT \cite{brown2020language}, LLaMA\cite{touvron2023llama1, touvron2023llama2, llama3modelcard}, and Gemini (formerly, Bard)\cite{team2023gemini}, which have been surprisingly successful in applications such as natural language processing (NLP) and tasks mimicking natural language understanding (NLU). These models are characterized by their vast number of parameters, deep learning capabilities, and extensive training data, which enable them to generate human-like text, comprehend complex instructions, and even perform creative tasks. 
In simple terms, LLMs are highly sophisticated autocomplete engines that learn to capture conditional probabilistic associations among data elements (called \textit{tokens}) at a massive scale. Although language models are an old idea from the 1980s, it is their recent avatar as LLMs that has caught everyone's attention. 

LLMs are based on a transformer architecture introduced by Vaswani et al. in their seminal paper \cite{vaswani2017attention}. This architecture enables the model to handle sequential data and understand the context in a more flexible and efficient manner compared to previous models, such as recurrent neural networks (RNNs) or long short-term memory (LSTM) networks. The training process involves unsupervised learning from a vast corpus of text data, allowing the model to grasp a wide range of language patterns, styles, and information. The massive scale of LLMs is \textit{the} defining feature. For example, GPT-3 (Generative Pre-trained Transformer-3)  developed by Open AI -- which is the model behind the chatbot ChatGPT -- has over 175 billion parameters and requires approximately {570 GB} of filtered text data to train. This immense scale allows the model to develop a broad and nuanced "understanding" of language and context, though it also raises challenges in terms of computational resources and potential biases in the training data. It turns out that what Nobel Laureate Philip Anderson said~\cite{anderson1972more} more than 50 years ago in the context of physics is valid in AI as well: "More is different", indeed!

LLMs seem to be good at certain applications, such as writing drafts, coding assistance, summarizing, translating, and answering questions. Despite their capabilities, LLMs face several limitations. They can generate misleading or biased information, relying on the data on which they were trained \cite{bender2021dangers}. Moreover, ethical considerations about consent, privacy, and misuse of generated content also plague these systems. This was most recently exemplified by the lawsuit filed by The New York Times against OpenAI and Microsoft for copyright infringement in December 2023, in which The New York Times claims that millions of articles were used to train AI systems, which are now viewed as a comparable source of information\cite{GrynbaumMacNYT2023}. In addition to such concerns, the sciences and engineering domains pose other challenges. LLMs are perhaps appropriate for domains such as NLP, where there are no conservation laws or first-principles-based knowledge to leverage. All relevant knowledge is in the data. Therefore, purely probability-based autocomplete techniques are appropriate and successful in these applications. 

However, domains such as chemical engineering are governed by the fundamental laws of physics and chemistry (and biology), constitutive relations, and highly technical knowledge about materials, processes, and systems\cite{venkatasubramanian2019promise, venkatasubramanian2022artificial}. Not using such a treasure trove of information seems not only inefficient, but also can lead to results that are unsafe. The cost of a mistake in movie and restaurant recommendations made using autocomplete-like guesses is not very high; maybe one loses a couple of hours and a few hundred dollars. However, guessing the wrong decision about the process could lead to potentially dangerous results. Although purely data-driven machine learning has immediate uses, we believe that the long-term success of AI in scientific and engineering domains would depend on the use of first principles and technical knowledge effectively. Current LLMs cannot reason or plan, as they lack such in-depth models of their domains. ChatGPT hallucinations are perhaps interesting, perhaps even amusing, in certain applications, but they can be potentially dangerous in highly technical domains such as chemical engineering. 

Furthermore, many chemical engineering applications are not "big data." We certainly have access to more data now than we did, say, a decade ago. But unlike vision, NLP, and game playing, we cannot easily generate terabytes of data, except perhaps in computer simulations. So, purely data-driven techniques mimicked from such domains are not appropriate. On the other hand, our knowledge of first principles can be leveraged and exploited to reduce the need for large amounts of data. Therefore, developing hybrid AI models is more appropriate for many chemical engineering applications\cite{venkatasubramanian2022artificial,mann2021predicting,mann2022hybrid,mann2023group,mann2023susie,mann2024esfiles,chakraborty2021ai,chakraborty2020mechanism}. The importance of using domain knowledge has become evident even in non-technical areas for LLMs. For example, ChatGPT uses human experts' guidance in the last stages of its training as reinforcement learning using human feedback (RLHF). Although this is a step in the right direction, much more needs to be done before the scientific and engineering domains can rely on these tools to solve problems. 

In this perspective article, we discuss the history of language models, progress in state-of-the-art systems, and their applications in a few areas. We also highlight the benefits accrued from the use of a hybrid AI approach ~\cite{chakraborty2022hybrid} in conjunction with LLMs for a few application domains. Finally, we conclude by suggesting some opportunities for work in the near term in this rapidly evolving field. 

\section{Evolution of Language Models}

The recent developments in the domain of large language models (LLMs) are groundbreaking in their impact on nearly every aspect of human life. To get a better sense of all this, it is useful to review the early stages of the advances that have enabled such astounding progress. In this section, we present the chronology of language models in three parts by splitting the timeline approximately based on the progress made, acknowledging the capabilities and limitations of that period. 

\subsection{Early Years: The Symbolic AI approach -- 1950-1990}\label{subsec:EarlyYears}

In the early part of AI's formative history, the predominant approach for knowledge modeling was based on \emph{symbolic logic}, such as the use of heuristic rules in expert systems, which yielded the name of that era as \emph{symbolic AI}. This is in contrast with recent developments that are more data-driven, \textit{i.e.}, \emph{numeric AI}, or machine learning (ML) as we know it today. 

The fact that language models are not novelties that have emerged only recently during the neural network phase of AI might seem surprising. The conceptualization of chatbot-like systems was made in the 1950s by the computing pioneer Alan Turing, as proposed in his groundbreaking article~\cite{TURINGTEST} describing the Turing test. This was quickly followed by the first forays into developing a chatbot, in the format we now know of, named \emph{ELIZA}~\cite{ELIZA}. Created nearly six decades ago, it is strikingly similar to the chatbots of today in many ways. Its key algorithm relies on scanning the user's input text for any \emph{keywords}, followed by presenting the output text according to a \emph{rule} related to the keyword(s) identified. If a keyword is not found, then either a context-free statement or a previous output is generated.

\begin{figure}
    \centering
    \includegraphics[width=\textwidth]{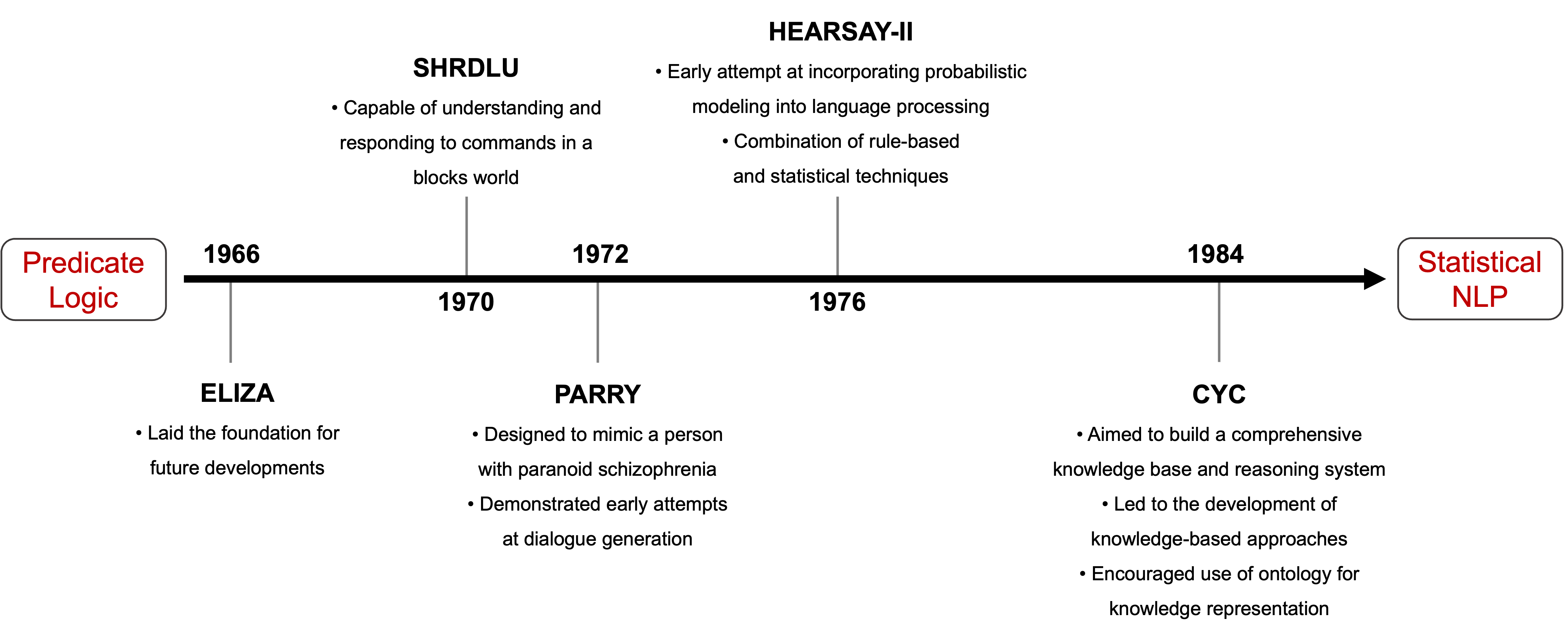}
    \caption{Progress of early language models, as research transitions from a logic-based approach to statistical-based approach.}
    \label{fig:earlydays}
\end{figure}

It is interesting to note that these rules are not specific to any task or subject and can be modified based on the end-user's application. ELIZA laid the foundation for the development of future chatbots by addressing challenges that must be overcome to advance the field. One such aspect highlighted was the need for such a conversational program to be capable of storing information obtained through its inputs. Subsequently, an information retrieval algorithm would be necessary to improve its conversational capabilities. In addition, such a tool should be able to make inferences based on the input received. This would enable it to interpret and store contextual information described in earlier input(s). These are characteristics that enable the state-of-the-art chatbots today to present themselves as human-like.

Although ELIZA presented the first conversational program, it had drawbacks such as its inability to parse the grammar and semantics of its language of use, lack of storage of information received during its input-output interactions, and absence of domain-specific information. In an attempt to address these shortcomings, SHRDLU~\cite{SHRDLU} was created in 1970. It relied on propositional logic, and semantic parsing of grammar rules to provide more flexibility to a conversational program in its learning. This would result in a more human-like interaction with the user. As a proof-of-concept, SHRDLU operated as if the user had been conversing with a one-eyed robot with a hand capable of interacting with blocks on a table. This interaction between the user, the robot, and the blocks in this environment demonstrates a deeper understanding of the subject than earlier systems.

Other systems based on the progress of that time were PARRY~\cite{PARRY}, a program that responded as if it were a paranoid schizophrenic. The output of the program was created to be a function of its input, beliefs, affects, and intentions~\cite{PARRY_criticisms}. It remains one of the first programs capable of dialogue generation. There have been intriguing conversations between PARRY (as the patient) and ELIZA (as the psychotherapist), which remain one of the few instances of interactions between computers in entirely natural language.

With the increasing success of previous language models, the Hearsay-II Speech Understanding System~\cite{HEARSAY2} was developed to convert speech to text that a computer program can communicate with. Its innovation came from a hierarchical structure in its organization of linguistic knowledge, which significantly facilitated information retrieval compared to its contemporaries. The system represented an early effort to integrate probabilistic modeling into language processing, utilizing a blend of rule-based and statistical techniques.

It is particularly interesting to note that efforts towards a hybrid approach towards language models and AI in general are not new. Several of the problems that are being addressed today were conceptually formalized in the 1980s and 1990s. Cyc~\cite{lenat2023getting}, a natural language system developed by Douglas Lenat nearly four decades ago, is one such tool. Today, it includes over 10 million logic rules coupled with a reasoning engine that is capable of making inferences. It uses a knowledge graph-like representation with more than 1,100 specialized reasoning modules. Dr. Lenat was once quoted as saying, "Intelligence is ten million rules." ~\cite{LENAT1991185, lenat1988} This is what his project, Cyc, has now acquired, and it is capable of combining human-like natural language understanding with complex scenarios.

\subsection{The data-driven statistical approach -- 1990-2017}\label{subsec:BeforeTransformer}

Despite the efforts of numerous researchers in NLP at the time, the complexities of natural language posed a barrier that required a radically different approach to overcome it, one that can learn from vast amounts of text without the need to be overly prescriptive in its design. With the availability of more data and computational power, largely due to Moore's law, data-driven statistical techniques made learning from textual sources considerably easier, and more computationally efficient. 

One of the first endeavors in statistical language modeling was the use of N-grams (the 1980s), which, combined with Hidden Markov Models (HMMs), led to considerable improvement in a variety of subdomains in NLP such as part-of-speech tagging, speech synthesis, and machine translation. The Viterbi algorithm~\cite{viterbi1967error} was adopted to decode the most likely sequence of hidden states in HMMs, improving the accuracy of NLP systems. These were followed by statistical methods, which combined information-theoretic approaches. These paved the way for the eventual application of ML-based approaches such as support vector machines (SVMs) in the early 2000s. Such improvements led to significant improvements in accuracy on a multitude of NLP tasks.

The growing number of datasets made available to the research community during this time further catalyzed the growth of statistics-based approaches for NLP tasks. The large-scale annotated corpora, Penn Treebank for English~\cite{PENN_TREEBANK}, revolutionized the field by providing researchers with standardized datasets to train and evaluate statistical models. These enabled researchers to explore more sophisticated statistical techniques. Soon, the earlier methods of logic-based and rule-based systems would not be able to scale up on these larger datasets.

With the rediscovery of the backpropagation algorithm in 1986~\cite{rumelhart1986learning}, the application of neural networks to natural language tasks began to take off. With the continued growth in computational speed and availability of more data than before, several application-specific variants of NNs were proposed. Recurrent neural networks (RNNs) were becoming popular because of their ability to understand the temporal aspect of data, making them ideal for processing sequential data as found in NLP. Improvements in conventional RNNs were the use of long-short-term-memory (LSTM) networks~\cite{hochreiter1997long, gers2001lstm}, which emerged as powerful models tailored for sequential data processing. The LSTM models addressed the vanishing gradient problem of RNNs, and were able to account for long-range dependencies, both drawbacks of the conventional RNN architecture. They soon became state-of-the-art models, surpassing their contemporary statistical model counterparts.

\subsection{The transformer revolution -- 2017-Present}\label{subsec:PostTransformer}

The introduction of the transformer architecture~\cite{vaswani2017attention} was a watershed moment for the field of NLP, which has since led to stunning progress. A major weakness of previous language models was the inability to retain information from the first elements of a sequence, which was lost when new elements were incorporated into the sequence. Instead of relying on the conventional practice of paying attention only to the last state of the encoder, as is typically done with RNNs, each step of the decoder in the transformer architecture examines all the states of the encoder. This approach enables the decoder to access information pertaining to all elements of the input sequence.

\begin{figure}
    \centering
    \includegraphics[width=0.7\textwidth]{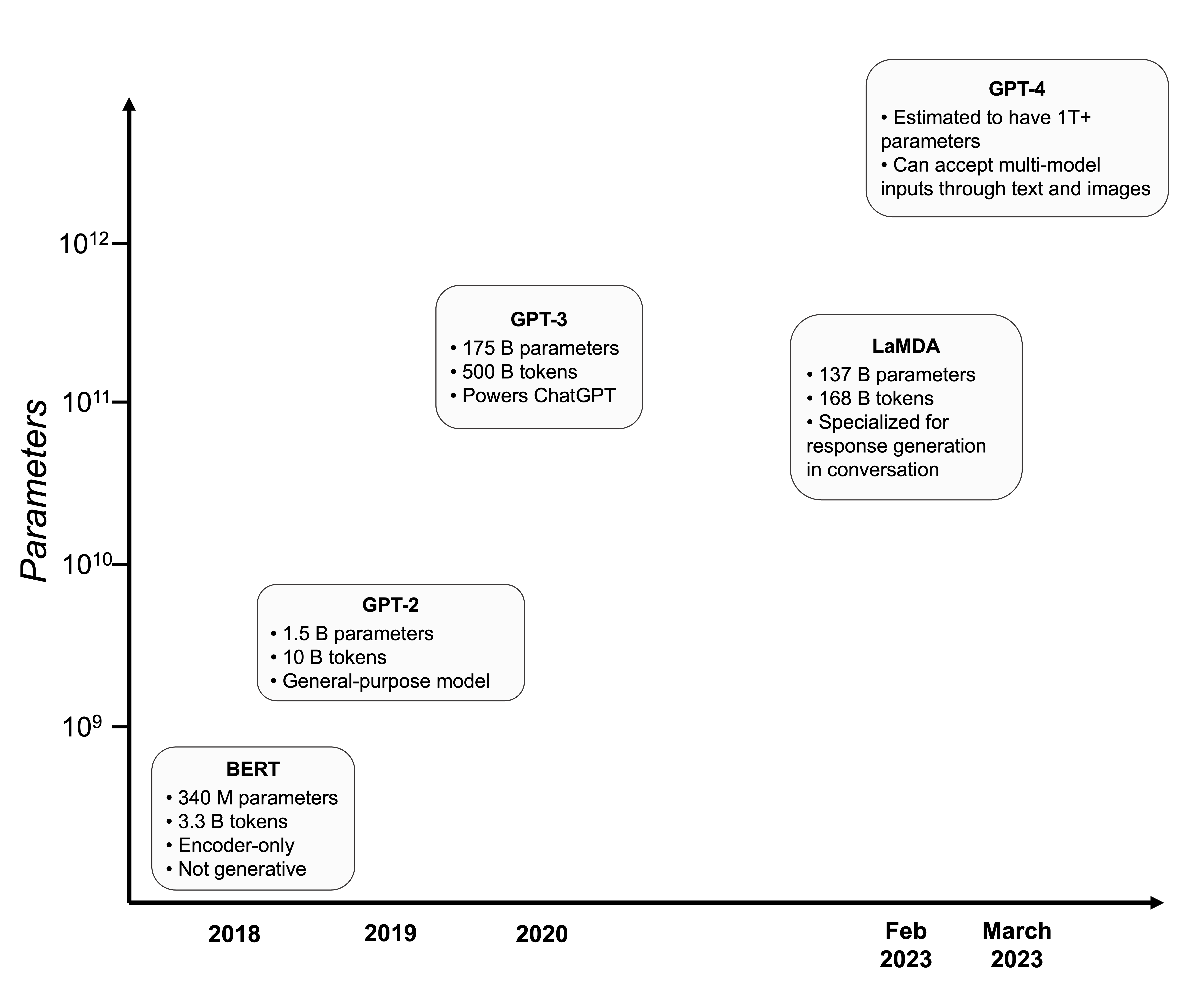}
    \caption{The rapid increase in the size of LLMs recently. Note the exponential increase in the number of parameters over a few years.}
    \label{fig:increasingDatasetSize&Params}
\end{figure}

This innovation revolutionized text processing and allowed language models to transition to \emph{large} language models (LLMs). This can be visualized with the increase in the size of the datasets used for training and the number of parameters in the model, depicted in Figure \ref{fig:increasingDatasetSize&Params}.

\subsubsection{ChatGPT}\label{subsec:ChatGPT}

In November 2022, OpenAI released ChatGPT, a chatbot that is capable of conversing with a user in a natural language similar to that of a human~\cite{brown2020language}. Additionally, it is able to follow instructions based on user prompts, and as a result, it can perform a wide range of tasks -- from writing prose and helping solve homework problems to writing code indistinguishable from that written by a professional programmer. It was a landmark moment for LLMs, since its meteoric rise to several million users in a matter of weeks trumped that of any social media platform in human history. This also opened several public discussions on the boons (and banes) of such technologies.

The key improvement in ChatGPT (alternatively, GPT-3/3.5), in contrast to its predecessors, is the fine-tuning approach applied to make ChatGPT more conversational in its usage. This fine-tuning procedure is based on a combination of supervised learning (with labeled data), and reinforcement learning with human feedback (RLHF)~\cite{christiano2017deep}.

\subsubsection{Training Data}

Since 2018, the number of parameters for each successive GPT model trained has increased exponentially, with a similar increase in the size of the training data. From the conventional ML perspective, this can be viewed as a way to combat the curse of dimensionality. Based on performance, the increase in parameters and training data has yielded more human-like characteristics to the outputs of these models.

\begin{table}[h!]
\centering
\begin{tabular}{c c c c} 
 \hline
 \textbf{Year} & \textbf{GPT version} & \textbf{Number of parameters (approximate)} & \textbf{Training corpus size (approximate)} \\ [0.5ex] 
 \hline
 % \hline
 2018 & GPT-1 & 117 M & 4.5 GB \\ 
 2019 & GPT-2 & 1.5 B & 40 GB \\
 2020 & GPT-3 & 175 B & 570 GB \\ [1ex] 
 \hline
\end{tabular}
\caption{The rapid increase in the size of training data used by GPT models by OpenAI. M -- million; B -- billion; GB -- gigabytes.}
\label{table:trainingdataforchatgpt}
\end{table}

Trained on a record-setting 570 GB of training data, GPT-3/3.5 is a successor to previous generative pre-trained transformer (GPT) models, namely GPT-1~\cite{radford2018improving} (2018), followed by GPT-2~\cite{radford2019language} (2019).

\subsubsection{Training Procedure} 

Training for ChatGPT is performed in 3 stages by building on the capabilities of InstructGPT~\cite{ouyang2022training}.

\paragraph{Stage 1: Generative Pre-training}

In the first stage, the model is trained on the training data using conventional NLP techniques such as masked language modeling (MLM), where a portion of the known output is \emph{masked}, and the goal of the trained model is to predict the same. As exemplified by BERT~\cite{devlin2018bert}, this approach creates a baseline model that is capable of text summarization, translation, and sentiment analysis tasks, among others.

\paragraph{Stage 2: Supervised fine-tuning}

Text summarization, translation, and sentiment analysis are one-shot tasks where an input is provided and the model produces an output. Beyond this task, the baseline model is fairly inept at having a conversation with a human on a particular topic, which was the goal of such a chatbot. As a result, it was necessary to fine-tune the baseline model into a supervised fine-tuned (SFT) model, for conversational purposes.

This second stage of training involved humans who created a supervised (labeled) dataset of expected (i.e., considered \emph{ideal}) responses for a set of inputs. These human labelers wrote an appropriate response to the input prompts, to which the baseline model was fine-tuned using stochastic gradient descent (SGD). This is a parameter optimization technique where the dataset is split into multiple smaller batches (mini-batch), and a mini-batch is randomly/stochastically chosen to calculate the gradient of the cost function. The parameters get updated subsequently, and this process is repeated until either a pre-specified value of the cost function is reached, or a pre-specified number of iterations have elapsed, whereby the optimization is terminated. Having obtained a SFT model, one still runs the risk of dealing with an \emph{overtrained} model, where the output merely reflects its memorization of the training dataset. This limitation could significantly narrow the scope of application, and diminish the model's efficacy, akin to that of a lookup table.

\paragraph{Stage 3: Reinforcement learning with human feedback (RLHF)}

In order to address this, a third and final stage of the training involved the use of reinforcement learning (RL), where the model's outputs are rewarded or penalized by \emph{learning} to provide outputs that maximize reward(s). The vanilla implementation of RL (without any modifications) would require a reward function -- which is not ideal for chatbots. Further, if such a mathematical function capable of predicting a better output (maximizing reward) existed, it would negate the additional steps being undertaken to fine-tune and improve the model.

Consequently, a human-in-the-loop approach was needed to develop a reward model. Here, human contractors were presented with four to nine outputs from the SFT model, and were asked to rank them from best to the worst responses for the given input prompt. The better the response, the higher the reward; therefore, the reward model learns which kind of response yields a higher reward.

Having discussed the reward model, it is important to update the SFT model so that it overcomes the limitations previously discussed in the earlier stage of training. Here, ChatGPT's model uses Proximal Policy Optimization (PPO)\cite{schulman2017proximal}, to update the \emph{policy} on the basis of which it returns its outputs. These updates are done in small steps so that the policy does not deviate significantly from what is considered a high-reward output. The information-theoretic concept of Kullback-Leibler (KL) divergence was used to measure such deviations, and to keep the ChatGPT model within reasonable bounds.

\subsubsection{Open-source LLMs}

\begin{table}[h!]
\centering
\caption{Open-source LLMs are not lagging behind their commercially successful counterparts. K -- thousand; M -- million; B -- billion; T -- trillion; GB -- gigabyte; TB -- terabyte.}
\begin{tabular}{|m{3em} m{7em} m{7em} m{10em} m{8em}|} 
 \hline
 \textbf{Year} & \textbf{GPT version} & \textbf{Number of parameters (approximate)} & \textbf{Training corpus size (approximate)} & \textbf{Number of tokens} \\ [0.5ex] 
 \hline \hline
 2022 & BLOOM~\cite{scao2022bloom} & 176 B & 1.6 TB & 350 B\\ 
 2023 & Llama~\cite{touvron2023llama1} & 65 B & - & 1.4 T\\
 2023 & Llama 2~\cite{touvron2023llama2} & 70 B & - & 2 T\\
 2023 & StableLM~\cite{StableLM} & 3 \& 7 B & 825 GB \cite{pile} & 1.5 T\\
 2023 & Pythia~\cite{biderman2023pythia} & 70 M - 12 B & 825 GB~\cite{pile} & 300-334 B\\
 2023 & Dolly 2.0~\cite{DatabricksBlog2023DollyV2} & 12 B & Fine-tuned on 15 K human-generated prompt / response pairs & Based on Pythia \cite{biderman2023pythia}\\
 2023 & Alpaca~\cite{alpaca} & 7 B & Fine-tuned on 52K instruction-following data & Fine-tuned from Llama 7B~\cite{touvron2023llama1}\\
 2023 & Vicuna~\cite{vicuna2023} & 13 B & Fine-tuned on 70K user-shared conversations gathered from ShareGPT.com with public APIs & Fine-tuned from Llama~\cite{touvron2023llama1}\\
 2024 & DBRX~\cite{dbrx} & 132 B & Pretrained on 12 T tokens of carefully curated data, and a maximum context length of 32k tokens & pretrained on 12 T tokens of text and code data\\
 2024 & Llama 3~\cite{llama3modelcard} & 8 B \& 80 B & - & 15 T\\[1ex]
 \hline
\end{tabular}
\label{table:opensourceLLMs}
\end{table}

The public release of ChatGPT led to a \emph{LLM race} between several AI research laboratories and commercial entities, resulting in the release of several open-source LLMs. These enable the public to tweak the parameters of the pre-trained models, which permit one to apply such a model's complex natural language capabilities to a specific domain.

The most popular open-source LLMs include BLOOM~\cite{scao2022bloom} and Llama~\cite{touvron2023llama1} by the technology giant Meta, which was quickly followed by Llama 2~\cite{touvron2023llama2}, and Llama 3~\cite{llama3modelcard}. Llama 2, having been trained on 2 trillion tokens, has double the context length of Llama 1. In addition, it offers 3 models of varying sizes (7B, 13B, and 70B parameters) so that the end user can utilize the model most appropriately sized for his or her task. Llama 2 has been trained using a similar approach to that of ChatGPT by using RLHF. The release of such LLMs has led to a wave of smaller, fine-tuned chatbots that perform comparably to their larger counterparts on specific tasks. These, and numerous other open-source LLMs/chatbots, have been presented in Table \ref{table:opensourceLLMs}.

\section{Modeling Knowledge in Chemical Engineering: The Three Paradigms}

Understanding the role of AI and LLMs in chemical engineering requires examining it through the lens of various knowledge-modeling approaches employed in our field. Initially, chemical engineering relied heavily on empirical methods and heuristics, with a conspicuous absence of quantitative, first-principles-based models for almost a century. This changed in the 1950s, marked by what is known as the Amundson era, in which the adoption of applied mathematical techniques, especially linear algebra, ordinary differential equations, and partial differential equations, revolutionized the field by enabling the creation of models for unit operations grounded in fundamental principles.

Similarly, decision-making within process systems engineering initially relied mainly on empirical knowledge and heuristic approaches. This paradigm shifted in the 1960s, marking another pivotal moment in modeling with the advent of mathematical programming techniques like mixed-integer linear programming (MILP) and mixed-integer nonlinear programming (MINLP). This transformation was pioneered by Roger Sargent, and his students. 

The next significant avatar in this long evolution of modeling paradigms is the introduction of knowledge representation concepts, and search techniques from artificial intelligence. This started in the early 1980s under the leadership of Westerberg, Stephanopoulos, and others from that era. After remaining in the background as a fringe activity for the past three decades, pursued by only a few researchers, this knowledge-modeling paradigm has now gone mainstream. The newest avatar in this evolution of knowledge models is the large-language model. While the classical AI models of the 1970s-1980s were based on symbolic representations using logic and reasoning, recent AI advances use statistical machine learning based on probability theory and network science. 

Broadly speaking, one might consider the Amundson era as the introduction of formal methods for modeling process units. The Sargent era, and the AI era are about modeling the process engineer. That is, modeling and automating human information processing and decision-making, formally, to solve problems in synthesis, design, control, scheduling, optimization, and risk analysis. Some of these could be addressed by the mathematical programming framework, \textit{i.e.}, the Sargent approach, but others, such as fault diagnosis and process hazards analysis, require causal model-based reasoning, and are better addressed by AI concepts and techniques. 

\subsection{Role of Symbolic AI in Knowledge Modeling and Representation}

When it comes to modeling knowledge, most chemical engineers immediately think of differential and algebraic equations (\textit{i.e.}, DAE models). These are suitable for certain classes of problems, such as those found in thermodynamics, transport phenomena, and reaction engineering. 

However, there are other kinds of knowledge that do not lend themselves to such models. For example, reasoning about cause and effect in a process plant is central to fault diagnosis, risk analysis, alarm management, and supervisory control. Knowledge modeling for this problem class typically does not lend itself to the traditional DAE approach to modeling because it cannot provide explicit relationships between cause(s) and effect(s). In some simple cases, perhaps it can, but it is incapable of addressing real-life industrial process systems, which are often complex and nonlinear with incomplete and/or uncertain data. Furthermore, even for simple systems, DAE-based models are not suitable for generating mechanistic explanations of causal behavior. This is where symbolic AI comes in.  

We should not forget that the conceptual breakthrough of representing and reasoning with symbolic structures and relationships is an essential contribution of AI\cite{rich1985artificial, venkatasubramanian1989cache, venkatasubramanian1990cache, russell2016artificial}. This is what we refer to as symbolic AI, the classical AI of the 1960s-1980s, to differentiate it from data-driven machine learning, which we call numeric AI. Although the importance of symbolic AI has been largely missed in all the current excitement about data-driven machine learning, we expect it to resurface as we go beyond purely data-driven models towards more comprehensive knowledge-based intelligent systems, which are necessary for many applications in chemical engineering.

The symbolic AI methodologies include models such as the following:

\begin{itemize}

\item Graph-theoretical models such as signed digraphs used extensively to perform causal reasoning in the identification of abnormal events, diagnosis, and risk analysis \cite{iri1979algorithm, vaidhyanathan1995digraph, maurya2003systematic, maurya2003systematic}
\item Petri nets used for modeling discrete event systems \cite{johnsson1998grafchart, viswanathan1998automatingpart1, viswanathan1998automatingpart2}
\item Rule-based production system models used in expert systems for automating higher-order reasoning \cite{banares1985knowledge, banares1985development, banares1987decade, banares1988decade, rich1987model, venkatasubramanian1988object}
\item Semantic network models such as ontologies used in materials discovery and design, domain-specific compilers, etc.\cite{banares1985knowledge, banares1985development, banares1987decade, banares1988decade, rich1987model, venkatasubramanian1988object, aldea2003ontology, venkatasubramanian2006ontological, marquardt2010re, hailemariam2010purdue1, hailemariam2010purdue2}
\item Object-oriented models such as agent-based models used in simulating the behavior and decision-making choices of independent, interacting, entities endowed with complex attributes and decision-making powers\cite{katare2001agent, julka2002agent}

\end{itemize}

All these have far-reaching consequences as we begin to develop more comprehensive hybrid-AI systems such as the following in the near future:
\begin{itemize}
\item Combining first-principles with data-driven processing\cite{sundaram2001design, viswanathan2002hybrid, ghosh2003sulfur, chakraborty2020mechanism, chakraborty2021ai}
\item Causal models-based explanatory systems\cite{vaidhyanathan1995digraph, rich1987model, venkatasubramanian1988object, rich1989causality, vedam1999pca, zhao2005phasuite1, zhao2005phasuite2}
\item Domain-specific knowledge engines\cite{venkatasubramanian2006ontological, caruthers2003catalyst, katare2004intelligent, hsu2008domain, suresh2010ontomodel1, suresh2010ontomodel2}
\end{itemize}

Thus, we don’t view AI methods as simply useful tools to extract patterns from large amounts of data, even though that benefit is very much there, but as a new knowledge modeling paradigm, the next natural evolutionary stage in the long history of developing formal methods: first applied math (i.e., differential and algebraic equations), then operations research (i.e., math programming) and now artificial intelligence. Conceptually, applied math models numerical relationships between variables and parameters, mathematical programming models relationships between constraints, and AI models relationships between symbolic variables and symbolic structures. In the early years, logic was considered the discipline best suited to provide the formal foundations of AI, but recent developments suggest that probability, statistics, and network science are perhaps better suited. The truth might lie in some combination of both, depending on the application. With generative AI poised to make a great impact in chemical engineering\cite{decardi2024generative}, increasing the accessibility of LLMs, application-specific training datasets, and the inclusion of expert domain knowledge into the model development pipeline, this is an exciting time for research in this field.

\section{Applications}\label{sec:Applications}

The science and engineering community has a myriad of possibilities to take advantage of this new and rapidly advancing technology. In this section, we discuss a few specific examples. It is not meant to be exhaustive, as the field is evolving rapidly, but is only suggestive of the new and exciting possibilities.

\subsection{Finetuning of LLMs}

The choice of LLMs for users is growing rapidly. Oftentimes, it is preferable to adapt an existing pre-trained LLM to a specific task. This \emph{fine-tuning} of LLMs has become exceedingly popular (and beneficial) in recent years. This fine-tuning is, in essence, transfer learning for LLMs. There are numerous benefits in fine-tuning LLMs: first, one can tailor the output of a general LLM to a specific domain of application; second, it reduces the computational load of training a much larger model with several billion parameters; third, it can reduce the number of trainable parameters for downstream tasks\cite{hu2021lora}.

LLMs have been shown to be capable of learning through successive prompts, and thus can learn in context\cite {wei2023larger}. Fine-tuning is a more \emph{permanent} exercise in contrast to in-context learning. Here, instead of tuning the trained parameters of the model, one meticulously crafts the inputs/prompts (prompt engineering) provided to the model. The objective is to guide the output to better harmonize with the desired outcome.

Building on the concept of in-concept learning is the idea of \emph{indexing}. This adds an information retrieval functionality to LLMs for extracting data from textual sources. There are numerous benefits accrued from in-concept learning, such as increased accuracy resulting in reduced hallucinations and semantic accuracy. The central theme of indexing LLMs is to use the model as a reasoning engine. Depending on the application domain, appropriate knowledge bases can be created, and semantic searches can be made on the same. This approach of combining a search engine with the natural language reasoning capabilities of an LLM was popularized by Lewis et al.\cite{lewis2020retrieval} as retrieval-augmented generation (RAG). A major benefit of such an approach is the ability to update the knowledge base with new information. In rapidly evolving fields such as drug discovery, biotechnology, and renewable energy, to name a few, it is vital that their databases be updated in real-time.

\subsection{Explanation Generation}

A major challenge for the sciences and engineering remains the explainability of mathematical models. The types of models obtained after thorough analyses are wide-ranging -- from white-box models (derived from first principles) to black-box models (\emph{e.g.,} neural networks), and gray-box models\cite{von2014hybrid, chakraborty2022hybrid} (which rely on combining the principles of both ends of the modeling spectrum). Although there are statistical approaches \cite{ribeiro2016should, lundberg2017unified} that attempt to explain these models, often for an end-user, an \emph{ideal} explanation would be one that is provided in a natural language fashion. In the domain of model discovery\cite{wilson2017alamo, chakraborty2020mechanism, udrescu2020ai, chakraborty2021ai, jul2023identifying, jul2024hybrid}, it is of prime importance to be able to explain the effects of the features, and their corresponding independent variables on the outputs.

In an attempt to take advantage of the human-like conversational capabilities of ChatGPT, we apply its prowess to generate meaningful sentences that give the illusion of \emph{understanding} in the same way as a human expert would, to the task of explaining mathematical equations and formulae. This is considerably more challenging than asking ChatGPT to write prose, or help with explaining a homework problem. This is because in order to explain the mathematical equation, a relevant knowledge-base is required for the retrieval of information pertinent to the query. Further, despite the abundance of training data provided, ChatGPT does not have information readily available regarding a new equation or mathematical relation that must be described in a natural language manner -- especially one that it has never seen during its training process.

Here, we emphasize the need (and success) of providing relevant context when utilizing such LLMs for generating an explanation in natural language. Let us consider a relatively simple mathematical equation that is well known in reaction kinetics, the Arrhenius equation: $k = A e^{-E_{a}/RT}$, where $k$ is the specific reaction rate constant, $E_{a}$ is the activation energy, $R$ is the universal gas constant, and $T$ is the temperature at which the reaction is occurring. For an expert in the field of reaction kinetics, the Arrhenius equation is an empirical relation denoting a negative-inverse exponential dependence on temperature along with an inverse-exponential dependence on the activation energy. Furthermore, such an expert can contrast this empirical relationship with another equation from the same field of study, \emph{e.g.,} the Eyring–Polanyi equation. Knowledge about the background, derivation, success and failure of the equation, details of the variables involved, etc, are essential for an explanation provided by a human.

For a capable program (such as ChatGPT) to produce human-like explanations for mathematical equations and relations, it will require (but is not limited to) the following:
\begin{enumerate}
    \item Information about the variables involved in the equation
    \item Background of the equation (\textit{e.g.,} its derivation)
    \item Any causal relationship(s) between the dependent and independent variables
    \item Additional dependencies of the parameters involved (\textit{e.g.,} temperature dependence of the specific rate constant as per the Arrhenius equation, in a rate law) 
\end{enumerate}

These details are typically found in the scientific literature, such as research articles, specialized textbooks on the subject matter, and industrial manuals, to name a few. The equation is often surrounded by additional metadata, such as references to related sources, equation numbers, derivations, etc. It is thus imperative that when either a human, or an LLM is asked to provide a human-like explanation for such an equation, the relevant context is accounted for since it has pertinent information that can be used effectively for the same.

\begin{figure}[h]
    \centering
    \includegraphics[width=0.9\textwidth]{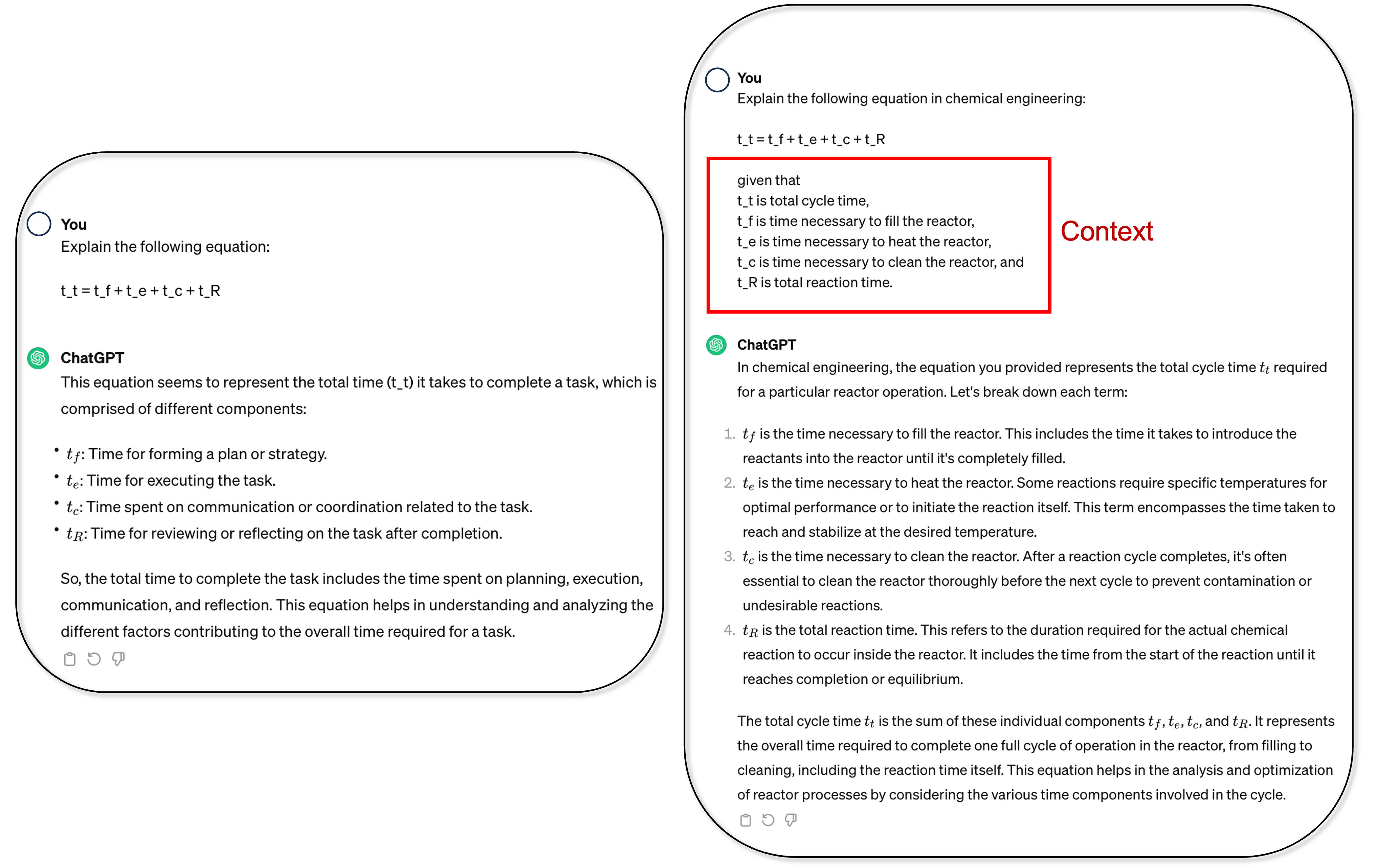}
    \caption{An example of how background context helps guide ChatGPT toward a better explanation of a simple equation from the domain of chemical reaction engineering. The additional context provided, is shown in the red colored box. This exemplifies the use of prompt engineering, which enables a more sensible output from LLMs, especially for more focused question-and-answering tasks such as the one highlighted here.}
    \label{fig:ExplanationGenExample}
\end{figure}

In conjunction with relevant context, we highlight the importance of prompt engineering for such an application. Pre-trained LLMs do not have all the relevant context for novel equations/mathematical relations that researchers may discover. Additionally, when a query is made to an LLM to describe the model, the relevant context must be provided first. This provides background information, on the basis of which the LLM can attempt to generate an explanation. One such example highlighting the benefit of added context is depicted in Figure \ref{fig:ExplanationGenExample}. It should be noted that all existing background information does not necessarily need to be provided at the outset. The user may choose to provide information in varying degrees of complexity, either following some existing guidelines provided in relevant textual source(s), or through the human expertise regarding the subject matter.

\subsection{LLMs in Chemistry}

The application of LLMs is not restricted to text-related tasks. In the chemical and biological sciences, many text-based representations exist that can be used to convert a conventional scientific problem into one that can be parsed in a textual format. The most popular choices are the Simplified Molecular Input Line Entry System (SMILES) \cite{weininger1988smiles, weininger1989smiles} to represent chemical structures, and FASTA\cite{lipman1985rapid, pearson1988improved} commonly used in bioinformatics. LLMs, with their textual reasoning capabilities, can be taken advantage of for such tasks. Guo et al.\cite{guo2023indeed} used a variety of GPT models for eight chemistry tasks. These include forward reaction and retrosynthetic prediction, among others, which have been predicted using domain-specific ML approaches. The results show its ability to perform well on text-related explanation tasks such as molecular captioning and text-based molecular design. Unsurprisingly, the results are not much better than the current state-of-the-art non-LLM approaches for the same tasks. This can be linked to an inherent lack of knowledge in the chemistry domain that would be essential -- say, for a human expert -- given the same tasks.

Consequently, an effort was made to enhance the capabilities of LLMs for chemistry-related tasks, titled ChemCrow\cite{bran2023chemcrow}, by integrating helper "tools" that work with the existing natural language reasoning capabilities and enable a wider range of actions. These include, but are not limited to, a web search tool, a literature search tool, a Python coding tool, a molecule manipulation tool (which can alter SMILES, and the corresponding molecules for subsequent tasks), etc. These improvements substantially improve the performance of the LLM and can be used to systematically obtain more relevant results for chemistry-related queries. Furthermore, this can allow the LLM to provide a reason for its steps taken toward obtaining the results, which can greatly benefit a human expert and can provide an additional avenue to explore. Although ChemCrow can have numerous advantages for the scientific community, it is necessary to have safety measures in place to reduce the risk of misuse of the same. Consequently, such tools were added to the framework. Further, for the safety of the user, the data is queried from the PubChem database and are provided in a simple manner to the user.

Similarly, graph neural nets that have traditionally been unable to use textual information and lack sufficient training data can be used in conjunction with language models for reaction prediction, as demonstrated by ReLM\cite{shi2023relm}. Here, reaction prediction is performed using a pre-trained language model and graph neural networks. Alternate approaches include using the capabilities of natural language in the design of organic structures, as was demonstrated by Ito et al.\cite{ito2024novo}. Here, GPT-4\cite{achiam2023gpt} was used as a natural language medium to obtain additional insights in an iterative manner.

\subsection{LLMs in biology, pharmaceuticals, and drug discovery}

\begin{figure}
    \centering
    \includegraphics[width=0.6\textwidth]{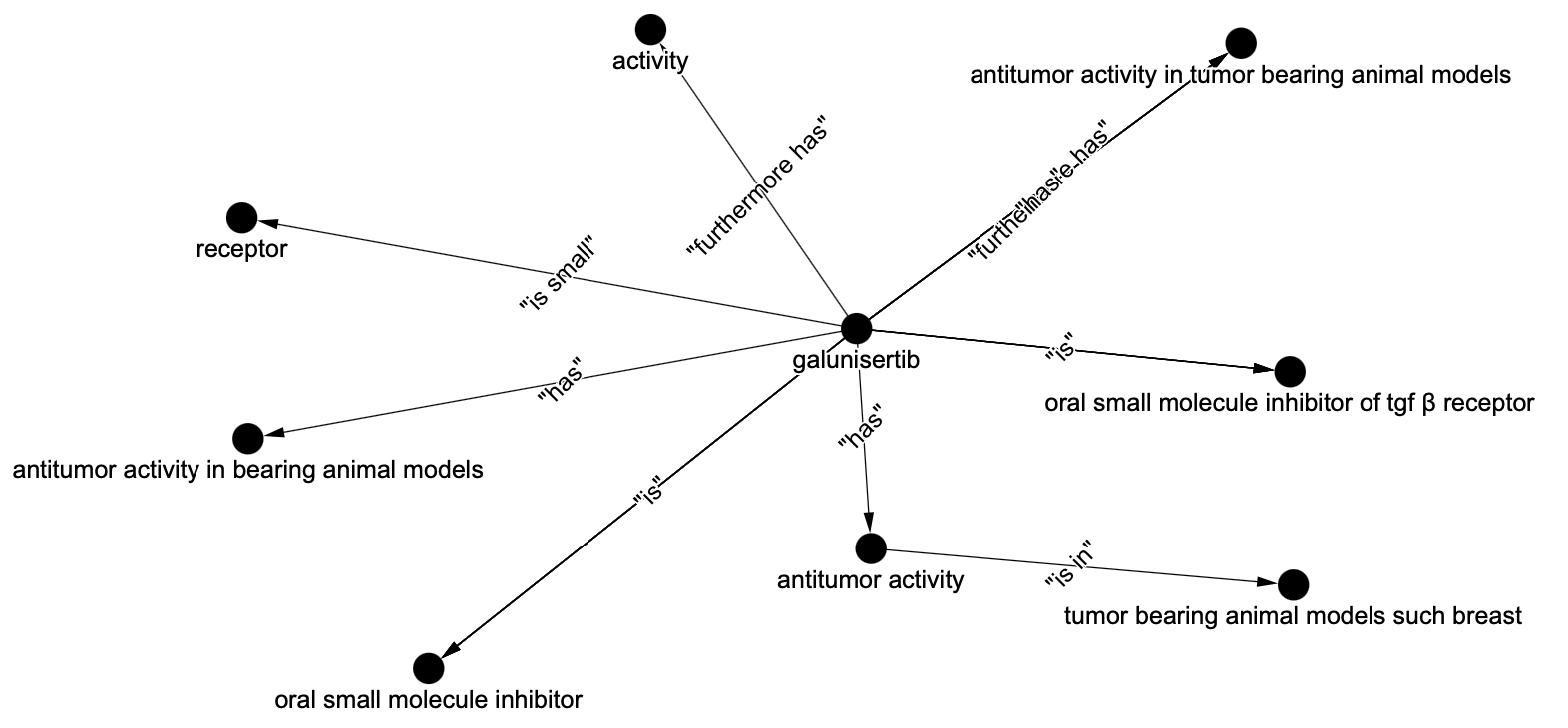}
    \caption{The output from SUSIE\cite{mann2023susie}, for an abstract in pharmaceutical sciences, discussing galunisertib\cite{herbertz2015clinical}.}
    \label{fig:susieExampleOutput}
\end{figure}

\begin{figure}
    \centering
    \includegraphics[width=0.5\textwidth]{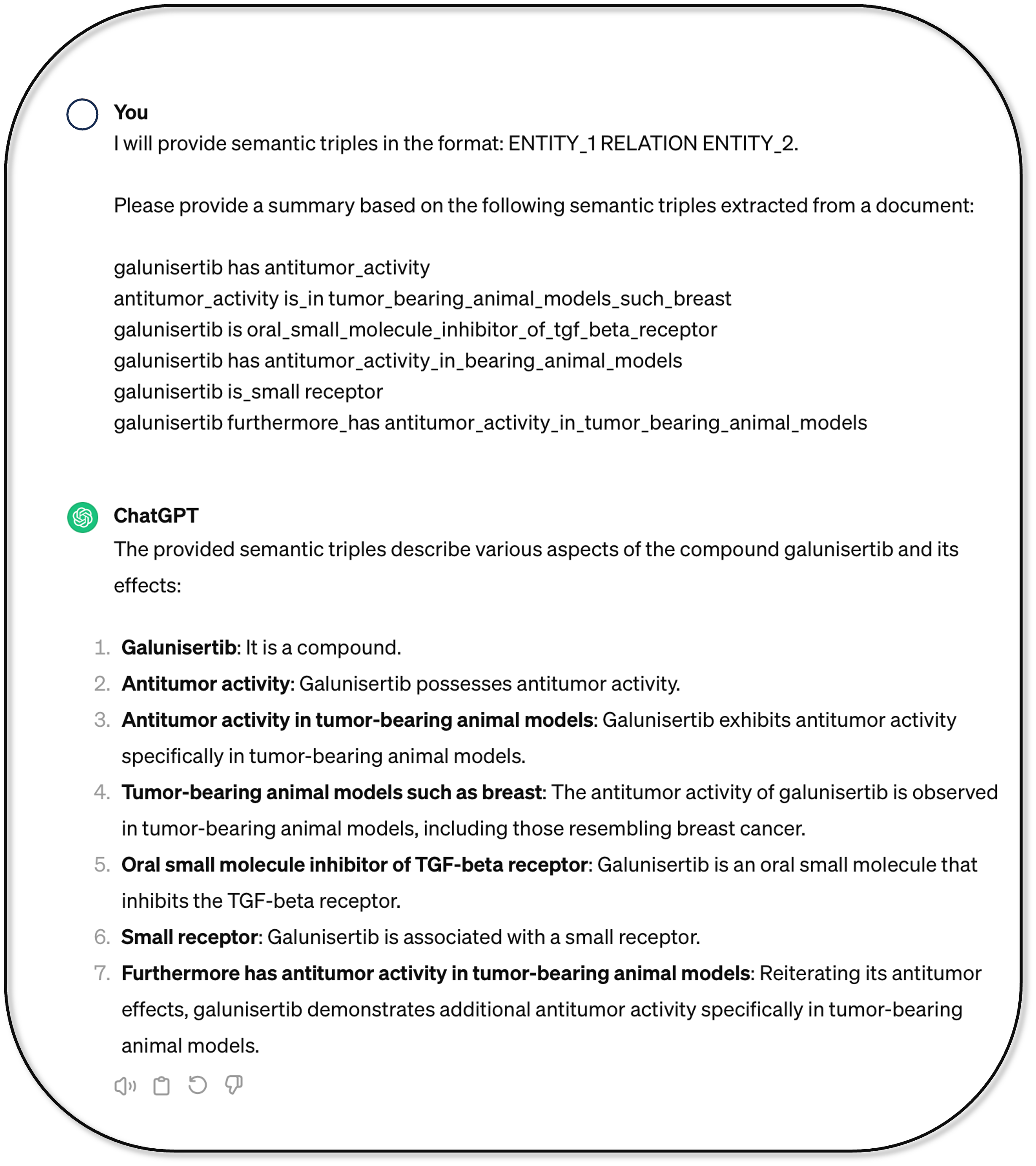}
    \caption{The output from ChatGPT for a summarization task when provided with a knowledge graph generated by  SUSIE\cite{mann2023susie}, for an abstract in pharmaceutical sciences, discussing galunisertib\cite{herbertz2015clinical}. The summarization of complex scientific text has become simpler, and more accurate by virtue of semantic triples extracted.}
    \label{fig:susieoutput2chatgpt}
\end{figure}

Large language models have also become useful in the pharmaceutical domain. A recent example of this is the extraction of semantic triples performed on pharmaceutical documents by Schema-based Unsupervised Semantic Information Extraction (SUSIE)\cite{mann2023susie}. Such a tool can accelerate drug discovery by searching for keywords and phrases through a corpus of documents in a fraction of the time that it would take a human expert. This system relied on fine-tuning a pre-trained domain-specific language model (BioBERT\cite{lee2020biobert}), on data from its intended application domain, followed by information extraction. Such a tool can drastically reduce the time taken by a human to undertake repetitive and tedious manual tasks, to mere seconds. A sample output from SUSIE based on an abstract of a journal article in the pharmaceutical sciences domain is depicted in Figure \ref{fig:susieExampleOutput}. When this output -- which is obtained after careful inclusion of domain-knowledge expertise into the data-driven pipeline -- is provided to an LLM (here, ChatGPT), it is capable of rendering an accurate natural language summary of the text. The same is shown in Figure \ref{fig:susieoutput2chatgpt}.

Other applications include ChatPathway\cite{li2023chatpathway}, which highlights the successful merging of domain knowledge in biology with the power of LLMs to predict biochemical reactions and their pathways. With the rapid increase in the number of applications of LLMs in the scientific and engineering domains, there has been a rise in scientific LLMs such as PubMedBERT\cite{gu2021domain}, MolGPT\cite{bagal2021molgpt}, SciBERT\cite{beltagy2019scibert} -- to name a few. For a more in-depth review of scientific LLMs, the reader is advised to refer to this comprehensive article by Zhang et al. \cite{zhang2024scientific} discussing the same in great detail.

\section{Next steps: From LLMs to LKMs}

With the advent of the LLMs, Pandora's box has been opened, revealing a multitude of exciting possibilities and serious concerns. Despite their impressive capabilities in certain applications, current LLMs have limited success in highly scientific and engineering applications. This is largely due to the lack of in-depth domain knowledge based on first principles. Current LLMs, in general, appear to be a mile wide in their scope but only a foot deep in fundamental principles. 

This limitation of purely data science models is due to their lack of “understanding” of the underlying knowledge. For example, a self-driving car can navigate impressively through traffic, but does it "know" and "understand" the concepts of mass, momentum, acceleration, force, and Newton's laws as we do? It does not. Its behavior is similar to that of a cheetah chasing an antelope in the wild. Both animals show great mastery of the dynamics of the chase, but do they “understand” these concepts? At best, current AI systems have perhaps achieved animal-like mastery of their tasks but have not gained a deeper "understanding" as many humans do. 

What we need are domain-specific LLMs that are perhaps a foot wide but a mile deep in domain knowledge. For example, a custom LLM in pharmaceutical engineering might not compose poems like Shakespeare, but it will not hallucinate on pharmaceutical chemistry. We need many such custom LLMs in different application domains that can communicate with one another.  

This brings up the importance of building first-principles knowledge in LLMs. For four decades, the first author has argued \cite{rich1987model, rich1989causality} about the importance of such hybrid AI models, combining mechanistic understanding based on the underlying physics, chemistry, and/or biology of our systems and processes with data-driven techniques. Our domain, in fact, most science and engineering, is governed by such fundamental principles. In this regard, it is different from areas such as game playing and computer vision, where there are no such conservation laws or constitutive equations to exploit. So, it makes sense in those domains to be mostly data-driven. But our applications are quite different. 

A related point is that many chemical engineering applications are not “big data.” We certainly have access to more data now than we did, say, a decade ago. But, unlike vision or game-playing, we don’t generate terabytes of data easily, except perhaps in computer simulations. So, purely data-driven techniques mimicked from such domains are not appropriate and might not work. On the other hand, our first-principles knowledge can be leveraged and exploited to reduce the need for large amounts of data. Thus, the construction of hybrid domain-specific LLMs is more appropriate for many chemical engineering applications. 

We classify AI applications into three categories \cite{venkatasubramanian2019artificial,vvyoutubechiang2019} – “easy,” “hard,” and “harder” problems. The relatively “easy” ones are those where plenty of data is available, and many off-the-shelf machine learning software could be used to extract useful patterns from such data. These are not particularly difficult problems as such things were demonstrated even in the 1990s. On the other hand, “hard” problems require the construction of hybrid AI models and the generation of mechanism-based causal explanations. Although work in the 1990s, again, showed~\cite{psichogios1992hybrid} how these could be done, we are still quite a ways from developing such models systematically, reliably, and easily for large-scale and diverse applications. We think this might take another five years or so. 

The third category, the “harder” problems, is those where one would have to build domain-specific ontologies, compilers, languages, etc., such as those reported for catalyst design \cite{caruthers2003catalyst, katare2004intelligent} and pharmaceutical manufacturing
\cite{venkatasubramanian2006ontological, hailemariam2010purdue1, hailemariam2010purdue2, mann2023susie}. Custom domain-specific LLMs fall into this category. This would require much more work, not the mere application of standard techniques and software in machine learning. They need careful integration of symbolic domain knowledge with data-driven methods. In fact, even generic LLMs such as ChatGPT have realized the importance of human experts and are already doing this under the guise of reinforcement learning with human feedback (RLHF). Another example of this realization is the development of \emph{AlphaGeometry}, which proves mathematical theorems at the Olympiad level~\cite{trinh2024solving}. As a neuro-symbolic system, it uses a language model in conjunction with a symbolic deduction engine. 

The domain-specific LLMs that we envision would require much more human expert guidance in the form of ontologies and heuristics, reflecting some of the techniques used in the expert systems era~\cite{venkatasubramanian2019promise}. Automation tools will certainly come along to make building such ontologies somewhat easier.

There are encouraging results on this front in recent work\cite{Roose2024NYT}. With the use of dictionary learning, researchers at Anthropic AI were able to investigate the underlying features\cite{templeton2024scaling} of their LLM named Claude Sonnet\cite{Roose2023NYT}. This is certainly a step in the right direction, as it enables researchers to better understand the shortcomings of these complex (to date, uninterpretable) models and aid in the future development of more explainable LLMs. Additional efforts at the frontiers of LLM research are geared towards developing small language models (SLMs), which are a fraction of the size of LLMs, are trained on higher quality data, and require significantly lower training computations than LLMs. One such example is that of Microsoft's Phi-1 SLM\cite{gunasekar2023textbooks}, which was trained on a textbook-quality dataset with a size of 7 billion tokens. Building on the success of the earlier model, Microsoft launched the Phi-3-mini SLM\cite{abdin2024phi}. These compact yet effective language models are more optimized versions of their larger predecessors.

Along these lines, an exciting, while nascent, avenue to explore is the inclusion of knowledge graphs (KGs) in an LLM framework. This enables one to circumvent the problems of hallucinations in LLMs\cite{ji2023survey}, and augment domain-specific knowledge to LLMs, resulting in a system that does not merely recall (occasionally, incorrect) factual information from its training data\cite{petroni2019language}. Due to the inherent lack of interpretability of an LLM, which functions as a black-box model due to its reasoning aided by a probability model\cite{zhang2022survey}, it is necessary to incorporate structured knowledge into the same, targeted for a specific domain. For a more comprehensive treatment of the use of structured knowledge representation in KGs along with enhanced inference and interpretability for LLMs, the reader is advised to refer to the review article by Pan et al.\cite{pan2023unifying}.

We call these hybrid AI systems \textit{Large Knowledge Models} (LKMs) because they will not be limited to NLP-based techniques or NLP-like applications only. We estimate that such systems might take about 5-10 years to emerge as routine implementations. In our opinion, the most interesting and intellectually challenging problems lie in the 'hard' and 'harder' categories. We think these are the areas where our PSE research community should focus now. 

In conclusion, generative AI has opened up unimagined possibilities in all aspects of human endeavor. It is poised to have a great impact on process systems engineering. However, to harness its potential safely and effectively, we need to go beyond large language models (LLMs) to large knowledge models (LKMs), which incorporate fundamental knowledge and human expertise deeply and effectively.

\section{Acknowledgements}
This paper is based on the plenary lecture delivered by the first author at the European Symposium of Computer-Aided Process Engineering (ESCAPE 33) in Athens, Greece, in June 2023. VV is grateful for the kind invitation and for the warm hospitality of Professor Antonis Kokossis and his colleagues at the conference. This work is supported in part by the NSF EFRI-DCheM 2132142 grant and funding from the Center for the Management of Systemic Risk (CMSR) at Columbia University.

\bibliographystyle{model1-num-names}
\bibliography{main.bib}

\begin{thebibliography}{122}
\expandafter\ifx\csname natexlab\endcsname\relax\def\natexlab#1{#1}\fi
\providecommand{\bibinfo}[2]{#2}
\ifx\xfnm\relax \def\xfnm[#1]{\unskip,\space#1}\fi
%Type = Article
\bibitem[{Brown et~al.(2020)Brown, Mann, Ryder, Subbiah, Kaplan, Dhariwal, Neelakantan, Shyam, Sastry, Askell et~al.}]{brown2020language}
\bibinfo{author}{T.~Brown}, \bibinfo{author}{B.~Mann}, \bibinfo{author}{N.~Ryder}, \bibinfo{author}{M.~Subbiah}, \bibinfo{author}{J.~D. Kaplan}, \bibinfo{author}{P.~Dhariwal}, \bibinfo{author}{A.~Neelakantan}, \bibinfo{author}{P.~Shyam}, \bibinfo{author}{G.~Sastry}, \bibinfo{author}{A.~Askell}, et~al.,
\newblock \bibinfo{title}{Language models are few-shot learners},
\newblock \bibinfo{journal}{Advances in neural information processing systems} \bibinfo{volume}{33} (\bibinfo{year}{2020}) \bibinfo{pages}{1877--1901}.
%Type = Article
\bibitem[{Touvron et~al.(2023{\natexlab{a}})Touvron, Lavril, Izacard, Martinet, Lachaux, Lacroix, Rozi{\`e}re, Goyal, Hambro, Azhar et~al.}]{touvron2023llama1}
\bibinfo{author}{H.~Touvron}, \bibinfo{author}{T.~Lavril}, \bibinfo{author}{G.~Izacard}, \bibinfo{author}{X.~Martinet}, \bibinfo{author}{M.-A. Lachaux}, \bibinfo{author}{T.~Lacroix}, \bibinfo{author}{B.~Rozi{\`e}re}, \bibinfo{author}{N.~Goyal}, \bibinfo{author}{E.~Hambro}, \bibinfo{author}{F.~Azhar}, et~al.,
\newblock \bibinfo{title}{Llama: Open and efficient foundation language models},
\newblock \bibinfo{journal}{arXiv preprint arXiv:2302.13971}  (\bibinfo{year}{2023}{\natexlab{a}}).
%Type = Article
\bibitem[{Touvron et~al.(2023{\natexlab{b}})Touvron, Martin, Stone, Albert, Almahairi, Babaei, Bashlykov, Batra, Bhargava, Bhosale et~al.}]{touvron2023llama2}
\bibinfo{author}{H.~Touvron}, \bibinfo{author}{L.~Martin}, \bibinfo{author}{K.~Stone}, \bibinfo{author}{P.~Albert}, \bibinfo{author}{A.~Almahairi}, \bibinfo{author}{Y.~Babaei}, \bibinfo{author}{N.~Bashlykov}, \bibinfo{author}{S.~Batra}, \bibinfo{author}{P.~Bhargava}, \bibinfo{author}{S.~Bhosale}, et~al.,
\newblock \bibinfo{title}{Llama 2: Open foundation and fine-tuned chat models},
\newblock \bibinfo{journal}{arXiv preprint arXiv:2307.09288}  (\bibinfo{year}{2023}{\natexlab{b}}).
%Type = Article
\bibitem[{AI@Meta(2024)}]{llama3modelcard}
\bibinfo{author}{AI@Meta},
\newblock \bibinfo{title}{Llama 3 model card}  (\bibinfo{year}{2024}).
%Type = Article
\bibitem[{Team et~al.(2023)Team, Anil, Borgeaud, Wu, Alayrac, Yu, Soricut, Schalkwyk, Dai, Hauth et~al.}]{team2023gemini}
\bibinfo{author}{G.~Team}, \bibinfo{author}{R.~Anil}, \bibinfo{author}{S.~Borgeaud}, \bibinfo{author}{Y.~Wu}, \bibinfo{author}{J.-B. Alayrac}, \bibinfo{author}{J.~Yu}, \bibinfo{author}{R.~Soricut}, \bibinfo{author}{J.~Schalkwyk}, \bibinfo{author}{A.~M. Dai}, \bibinfo{author}{A.~Hauth}, et~al.,
\newblock \bibinfo{title}{Gemini: a family of highly capable multimodal models},
\newblock \bibinfo{journal}{arXiv preprint arXiv:2312.11805}  (\bibinfo{year}{2023}).
%Type = Article
\bibitem[{Vaswani et~al.(2017)Vaswani, Shazeer, Parmar, Uszkoreit, Jones, Gomez, Kaiser, and Polosukhin}]{vaswani2017attention}
\bibinfo{author}{A.~Vaswani}, \bibinfo{author}{N.~Shazeer}, \bibinfo{author}{N.~Parmar}, \bibinfo{author}{J.~Uszkoreit}, \bibinfo{author}{L.~Jones}, \bibinfo{author}{A.~N. Gomez}, \bibinfo{author}{{\L}.~Kaiser}, \bibinfo{author}{I.~Polosukhin},
\newblock \bibinfo{title}{Attention is all you need},
\newblock \bibinfo{journal}{Advances in neural information processing systems} \bibinfo{volume}{30} (\bibinfo{year}{2017}).
%Type = Article
\bibitem[{Anderson(1972)}]{anderson1972more}
\bibinfo{author}{P.~W. Anderson},
\newblock \bibinfo{title}{More is different},
\newblock \bibinfo{journal}{Science} \bibinfo{volume}{177} (\bibinfo{year}{1972}) \bibinfo{pages}{393--396}.
%Type = Inproceedings
\bibitem[{Bender et~al.(2021)Bender, Gebru, McMillan-Major, and Shmitchell}]{bender2021dangers}
\bibinfo{author}{E.~M. Bender}, \bibinfo{author}{T.~Gebru}, \bibinfo{author}{A.~McMillan-Major}, \bibinfo{author}{S.~Shmitchell},
\newblock \bibinfo{title}{On the dangers of stochastic parrots: Can language models be too big?},
\newblock in: \bibinfo{booktitle}{Proceedings of the 2021 ACM conference on fairness, accountability, and transparency}, pp. \bibinfo{pages}{610--623}.
%Type = Article
\bibitem[{Grynbaum and Mac(2024)}]{GrynbaumMacNYT2023}
\bibinfo{author}{M.~M. Grynbaum}, \bibinfo{author}{R.~Mac},
\newblock \bibinfo{title}{The times sues openai and microsoft over a.i. use of copyrighted work},
\newblock \bibinfo{journal}{The New York Times}  (\bibinfo{year}{2024}).
%Type = Article
\bibitem[{Venkatasubramanian(2019)}]{venkatasubramanian2019promise}
\bibinfo{author}{V.~Venkatasubramanian},
\newblock \bibinfo{title}{The promise of artificial intelligence in chemical engineering: Is it here, finally?},
\newblock \bibinfo{journal}{AIChE Journal} \bibinfo{volume}{65} (\bibinfo{year}{2019}) \bibinfo{pages}{466--478}.
%Type = Article
\bibitem[{Venkatasubramanian and Mann(2022)}]{venkatasubramanian2022artificial}
\bibinfo{author}{V.~Venkatasubramanian}, \bibinfo{author}{V.~Mann},
\newblock \bibinfo{title}{Artificial intelligence in reaction prediction and chemical synthesis},
\newblock \bibinfo{journal}{Current Opinion in Chemical Engineering} \bibinfo{volume}{36} (\bibinfo{year}{2022}) \bibinfo{pages}{100749}.
%Type = Article
\bibitem[{Mann and Venkatasubramanian(2021)}]{mann2021predicting}
\bibinfo{author}{V.~Mann}, \bibinfo{author}{V.~Venkatasubramanian},
\newblock \bibinfo{title}{Predicting chemical reaction outcomes: A grammar ontology-based transformer framework},
\newblock \bibinfo{journal}{AIChE Journal} \bibinfo{volume}{67} (\bibinfo{year}{2021}) \bibinfo{pages}{e17190}.
%Type = Article
\bibitem[{Mann et~al.(2022)Mann, Brito, Gani, and Venkatasubramanian}]{mann2022hybrid}
\bibinfo{author}{V.~Mann}, \bibinfo{author}{K.~Brito}, \bibinfo{author}{R.~Gani}, \bibinfo{author}{V.~Venkatasubramanian},
\newblock \bibinfo{title}{Hybrid, interpretable machine learning for thermodynamic property estimation using grammar2vec for molecular representation},
\newblock \bibinfo{journal}{Fluid Phase Equilibria} \bibinfo{volume}{561} (\bibinfo{year}{2022}) \bibinfo{pages}{113531}.
%Type = Article
\bibitem[{Mann et~al.(2023{\natexlab{a}})Mann, Gani, and Venkatasubramanian}]{mann2023group}
\bibinfo{author}{V.~Mann}, \bibinfo{author}{R.~Gani}, \bibinfo{author}{V.~Venkatasubramanian},
\newblock \bibinfo{title}{Group contribution-based property modeling for chemical product design: A perspective in the ai era},
\newblock \bibinfo{journal}{Fluid Phase Equilibria} \bibinfo{volume}{568} (\bibinfo{year}{2023}{\natexlab{a}}) \bibinfo{pages}{113734}.
%Type = Article
\bibitem[{Mann et~al.(2023{\natexlab{b}})Mann, Viswanath, Vaidyaraman, Balakrishnan, and Venkatasubramanian}]{mann2023susie}
\bibinfo{author}{V.~Mann}, \bibinfo{author}{S.~Viswanath}, \bibinfo{author}{S.~Vaidyaraman}, \bibinfo{author}{J.~Balakrishnan}, \bibinfo{author}{V.~Venkatasubramanian},
\newblock \bibinfo{title}{Susie: Pharmaceutical cmc ontology-based information extraction for drug development using machine learning},
\newblock \bibinfo{journal}{Computers \& Chemical Engineering} \bibinfo{volume}{179} (\bibinfo{year}{2023}{\natexlab{b}}) \bibinfo{pages}{108446}.
%Type = Article
\bibitem[{Mann et~al.(2024)Mann, Sales-Cruz, Gani, and Venkatasubramanian}]{mann2024esfiles}
\bibinfo{author}{V.~Mann}, \bibinfo{author}{M.~Sales-Cruz}, \bibinfo{author}{R.~Gani}, \bibinfo{author}{V.~Venkatasubramanian},
\newblock \bibinfo{title}{esfiles: Intelligent process flowsheet synthesis using process knowledge, symbolic ai, and machine learning},
\newblock \bibinfo{journal}{Computers \& Chemical Engineering} \bibinfo{volume}{181} (\bibinfo{year}{2024}) \bibinfo{pages}{108505}.
%Type = Article
\bibitem[{Chakraborty et~al.(2021)Chakraborty, Sivaram, and Venkatasubramanian}]{chakraborty2021ai}
\bibinfo{author}{A.~Chakraborty}, \bibinfo{author}{A.~Sivaram}, \bibinfo{author}{V.~Venkatasubramanian},
\newblock \bibinfo{title}{Ai-darwin: A first principles-based model discovery engine using machine learning},
\newblock \bibinfo{journal}{Computers \& Chemical Engineering} \bibinfo{volume}{154} (\bibinfo{year}{2021}) \bibinfo{pages}{107470}.
%Type = Article
\bibitem[{Chakraborty et~al.(2020)Chakraborty, Sivaram, Samavedham, and Venkatasubramanian}]{chakraborty2020mechanism}
\bibinfo{author}{A.~Chakraborty}, \bibinfo{author}{A.~Sivaram}, \bibinfo{author}{L.~Samavedham}, \bibinfo{author}{V.~Venkatasubramanian},
\newblock \bibinfo{title}{Mechanism discovery and model identification using genetic feature extraction and statistical testing},
\newblock \bibinfo{journal}{Computers \& Chemical Engineering} \bibinfo{volume}{140} (\bibinfo{year}{2020}) \bibinfo{pages}{106900}.
%Type = Article
\bibitem[{Chakraborty et~al.(2022)Chakraborty, Serneels, Claussen, and Venkatasubramanian}]{chakraborty2022hybrid}
\bibinfo{author}{A.~Chakraborty}, \bibinfo{author}{S.~Serneels}, \bibinfo{author}{H.~Claussen}, \bibinfo{author}{V.~Venkatasubramanian},
\newblock \bibinfo{title}{Hybrid ai models in chemical engineering--a purpose-driven perspective},
\newblock \bibinfo{journal}{Computer Aided Chemical Engineering} \bibinfo{volume}{51} (\bibinfo{year}{2022}) \bibinfo{pages}{1507--1512}.
%Type = Article
\bibitem[{Turing(1950)}]{TURINGTEST}
\bibinfo{author}{A.~M. Turing},
\newblock \bibinfo{title}{{Computing Machinery And Intelligence}},
\newblock \bibinfo{journal}{Mind} \bibinfo{volume}{LIX} (\bibinfo{year}{1950}) \bibinfo{pages}{433--460}.
%Type = Article
\bibitem[{Weizenbaum(1966)}]{ELIZA}
\bibinfo{author}{J.~Weizenbaum},
\newblock \bibinfo{title}{Eliza—a computer program for the study of natural language communication between man and machine},
\newblock \bibinfo{journal}{Communications of the ACM} \bibinfo{volume}{9} (\bibinfo{year}{1966}) \bibinfo{pages}{36--45}.
%Type = Techreport
\bibitem[{Winograd(1971)}]{SHRDLU}
\bibinfo{author}{T.~Winograd}, \bibinfo{title}{Procedures as a representation for data in a computer program for understanding natural language}, \bibinfo{type}{Technical Report}, MASSACHUSETTS INST OF TECH CAMBRIDGE PROJECT MAC, \bibinfo{year}{1971}.
%Type = Article
\bibitem[{Colby et~al.(1971)Colby, Weber, and Hilf}]{PARRY}
\bibinfo{author}{K.~M. Colby}, \bibinfo{author}{S.~Weber}, \bibinfo{author}{F.~D. Hilf},
\newblock \bibinfo{title}{Artificial paranoia},
\newblock \bibinfo{journal}{Artificial intelligence} \bibinfo{volume}{2} (\bibinfo{year}{1971}) \bibinfo{pages}{1--25}.
%Type = Article
\bibitem[{Colby(1974)}]{PARRY_criticisms}
\bibinfo{author}{K.~M. Colby},
\newblock \bibinfo{title}{Ten criticisms of parry},
\newblock \bibinfo{journal}{ACM SIGART Bulletin}  (\bibinfo{year}{1974}) \bibinfo{pages}{5--9}.
%Type = Article
\bibitem[{Erman et~al.(1980)Erman, Hayes-Roth, Lesser, and Reddy}]{HEARSAY2}
\bibinfo{author}{L.~D. Erman}, \bibinfo{author}{F.~Hayes-Roth}, \bibinfo{author}{V.~R. Lesser}, \bibinfo{author}{D.~R. Reddy},
\newblock \bibinfo{title}{The hearsay-ii speech-understanding system: Integrating knowledge to resolve uncertainty},
\newblock \bibinfo{journal}{ACM Computing Surveys (CSUR)} \bibinfo{volume}{12} (\bibinfo{year}{1980}) \bibinfo{pages}{213--253}.
%Type = Article
\bibitem[{Lenat and Marcus(2023)}]{lenat2023getting}
\bibinfo{author}{D.~Lenat}, \bibinfo{author}{G.~Marcus},
\newblock \bibinfo{title}{Getting from generative ai to trustworthy ai: What llms might learn from cyc},
\newblock \bibinfo{journal}{arXiv preprint arXiv:2308.04445}  (\bibinfo{year}{2023}).
%Type = Article
\bibitem[{Lenat and Feigenbaum(1991)}]{LENAT1991185}
\bibinfo{author}{D.~B. Lenat}, \bibinfo{author}{E.~A. Feigenbaum},
\newblock \bibinfo{title}{On the thresholds of knowledge},
\newblock \bibinfo{journal}{Artificial Intelligence} \bibinfo{volume}{47} (\bibinfo{year}{1991}) \bibinfo{pages}{185--250}.
%Type = Inproceedings
\bibitem[{Lenat(1988)}]{lenat1988}
\bibinfo{author}{D.~B. Lenat},
\newblock \bibinfo{title}{The case for inelegance},
\newblock in: \bibinfo{booktitle}{Proceedings of the International Workshop on Artificial Intelligence for Industrial Applications, Tokyo}.
%Type = Article
\bibitem[{Viterbi(1967)}]{viterbi1967error}
\bibinfo{author}{A.~Viterbi},
\newblock \bibinfo{title}{Error bounds for convolutional codes and an asymptotically optimum decoding algorithm},
\newblock \bibinfo{journal}{IEEE transactions on Information Theory} \bibinfo{volume}{13} (\bibinfo{year}{1967}) \bibinfo{pages}{260--269}.
%Type = Article
\bibitem[{Marcus et~al.(1993)Marcus, Santorini, and Marcinkiewicz}]{PENN_TREEBANK}
\bibinfo{author}{M.~Marcus}, \bibinfo{author}{B.~Santorini}, \bibinfo{author}{M.~A. Marcinkiewicz},
\newblock \bibinfo{title}{Building a large annotated corpus of english: The penn treebank},
\newblock \bibinfo{journal}{Computational linguistics} \bibinfo{volume}{19} (\bibinfo{year}{1993}) \bibinfo{pages}{313--330}.
%Type = Article
\bibitem[{Rumelhart et~al.(1986)Rumelhart, Hinton, and Williams}]{rumelhart1986learning}
\bibinfo{author}{D.~E. Rumelhart}, \bibinfo{author}{G.~E. Hinton}, \bibinfo{author}{R.~J. Williams},
\newblock \bibinfo{title}{Learning representations by back-propagating errors},
\newblock \bibinfo{journal}{nature} \bibinfo{volume}{323} (\bibinfo{year}{1986}) \bibinfo{pages}{533--536}.
%Type = Article
\bibitem[{Hochreiter and Schmidhuber(1997)}]{hochreiter1997long}
\bibinfo{author}{S.~Hochreiter}, \bibinfo{author}{J.~Schmidhuber},
\newblock \bibinfo{title}{Long short-term memory},
\newblock \bibinfo{journal}{Neural computation} \bibinfo{volume}{9} (\bibinfo{year}{1997}) \bibinfo{pages}{1735--1780}.
%Type = Article
\bibitem[{Gers and Schmidhuber(2001)}]{gers2001lstm}
\bibinfo{author}{F.~A. Gers}, \bibinfo{author}{E.~Schmidhuber},
\newblock \bibinfo{title}{Lstm recurrent networks learn simple context-free and context-sensitive languages},
\newblock \bibinfo{journal}{IEEE transactions on neural networks} \bibinfo{volume}{12} (\bibinfo{year}{2001}) \bibinfo{pages}{1333--1340}.
%Type = Article
\bibitem[{Christiano et~al.(2017)Christiano, Leike, Brown, Martic, Legg, and Amodei}]{christiano2017deep}
\bibinfo{author}{P.~F. Christiano}, \bibinfo{author}{J.~Leike}, \bibinfo{author}{T.~Brown}, \bibinfo{author}{M.~Martic}, \bibinfo{author}{S.~Legg}, \bibinfo{author}{D.~Amodei},
\newblock \bibinfo{title}{Deep reinforcement learning from human preferences},
\newblock \bibinfo{journal}{Advances in neural information processing systems} \bibinfo{volume}{30} (\bibinfo{year}{2017}).
%Type = Article
\bibitem[{Radford et~al.(2018)Radford, Narasimhan, Salimans, Sutskever et~al.}]{radford2018improving}
\bibinfo{author}{A.~Radford}, \bibinfo{author}{K.~Narasimhan}, \bibinfo{author}{T.~Salimans}, \bibinfo{author}{I.~Sutskever}, et~al.,
\newblock \bibinfo{title}{Improving language understanding by generative pre-training}  (\bibinfo{year}{2018}).
%Type = Article
\bibitem[{Radford et~al.(2019)Radford, Wu, Child, Luan, Amodei, Sutskever et~al.}]{radford2019language}
\bibinfo{author}{A.~Radford}, \bibinfo{author}{J.~Wu}, \bibinfo{author}{R.~Child}, \bibinfo{author}{D.~Luan}, \bibinfo{author}{D.~Amodei}, \bibinfo{author}{I.~Sutskever}, et~al.,
\newblock \bibinfo{title}{Language models are unsupervised multitask learners},
\newblock \bibinfo{journal}{OpenAI blog} \bibinfo{volume}{1} (\bibinfo{year}{2019}) \bibinfo{pages}{9}.
%Type = Article
\bibitem[{Ouyang et~al.(2022)Ouyang, Wu, Jiang, Almeida, Wainwright, Mishkin, Zhang, Agarwal, Slama, Ray et~al.}]{ouyang2022training}
\bibinfo{author}{L.~Ouyang}, \bibinfo{author}{J.~Wu}, \bibinfo{author}{X.~Jiang}, \bibinfo{author}{D.~Almeida}, \bibinfo{author}{C.~Wainwright}, \bibinfo{author}{P.~Mishkin}, \bibinfo{author}{C.~Zhang}, \bibinfo{author}{S.~Agarwal}, \bibinfo{author}{K.~Slama}, \bibinfo{author}{A.~Ray}, et~al.,
\newblock \bibinfo{title}{Training language models to follow instructions with human feedback},
\newblock \bibinfo{journal}{Advances in Neural Information Processing Systems} \bibinfo{volume}{35} (\bibinfo{year}{2022}) \bibinfo{pages}{27730--27744}.
%Type = Article
\bibitem[{Devlin et~al.(2018)Devlin, Chang, Lee, and Toutanova}]{devlin2018bert}
\bibinfo{author}{J.~Devlin}, \bibinfo{author}{M.-W. Chang}, \bibinfo{author}{K.~Lee}, \bibinfo{author}{K.~Toutanova},
\newblock \bibinfo{title}{Bert: Pre-training of deep bidirectional transformers for language understanding},
\newblock \bibinfo{journal}{arXiv preprint arXiv:1810.04805}  (\bibinfo{year}{2018}).
%Type = Article
\bibitem[{Schulman et~al.(2017)Schulman, Wolski, Dhariwal, Radford, and Klimov}]{schulman2017proximal}
\bibinfo{author}{J.~Schulman}, \bibinfo{author}{F.~Wolski}, \bibinfo{author}{P.~Dhariwal}, \bibinfo{author}{A.~Radford}, \bibinfo{author}{O.~Klimov},
\newblock \bibinfo{title}{Proximal policy optimization algorithms},
\newblock \bibinfo{journal}{arXiv preprint arXiv:1707.06347}  (\bibinfo{year}{2017}).
%Type = Article
\bibitem[{Scao et~al.(2022)Scao, Fan, Akiki, Pavlick, Ili{\'c}, Hesslow, Castagn{\'e}, Luccioni, Yvon, Gall{\'e} et~al.}]{scao2022bloom}
\bibinfo{author}{T.~L. Scao}, \bibinfo{author}{A.~Fan}, \bibinfo{author}{C.~Akiki}, \bibinfo{author}{E.~Pavlick}, \bibinfo{author}{S.~Ili{\'c}}, \bibinfo{author}{D.~Hesslow}, \bibinfo{author}{R.~Castagn{\'e}}, \bibinfo{author}{A.~S. Luccioni}, \bibinfo{author}{F.~Yvon}, \bibinfo{author}{M.~Gall{\'e}}, et~al.,
\newblock \bibinfo{title}{Bloom: A 176b-parameter open-access multilingual language model},
\newblock \bibinfo{journal}{arXiv preprint arXiv:2211.05100}  (\bibinfo{year}{2022}).
%Type = Misc
\bibitem[{AI(2023)}]{StableLM}
\bibinfo{author}{S.~AI}, \bibinfo{title}{Stability ai launches the first of its stablelm suite of language models}, \bibinfo{year}{2023}.
%Type = Article
\bibitem[{Gao et~al.(2020)Gao, Biderman, Black, Golding, Hoppe, Foster, Phang, He, Thite, Nabeshima, Presser, and Leahy}]{pile}
\bibinfo{author}{L.~Gao}, \bibinfo{author}{S.~Biderman}, \bibinfo{author}{S.~Black}, \bibinfo{author}{L.~Golding}, \bibinfo{author}{T.~Hoppe}, \bibinfo{author}{C.~Foster}, \bibinfo{author}{J.~Phang}, \bibinfo{author}{H.~He}, \bibinfo{author}{A.~Thite}, \bibinfo{author}{N.~Nabeshima}, \bibinfo{author}{S.~Presser}, \bibinfo{author}{C.~Leahy},
\newblock \bibinfo{title}{The {P}ile: An 800gb dataset of diverse text for language modeling},
\newblock \bibinfo{journal}{arXiv preprint arXiv:2101.00027}  (\bibinfo{year}{2020}).
%Type = Inproceedings
\bibitem[{Biderman et~al.(2023)Biderman, Schoelkopf, Anthony, Bradley, O’Brien, Hallahan, Khan, Purohit, Prashanth, Raff et~al.}]{biderman2023pythia}
\bibinfo{author}{S.~Biderman}, \bibinfo{author}{H.~Schoelkopf}, \bibinfo{author}{Q.~G. Anthony}, \bibinfo{author}{H.~Bradley}, \bibinfo{author}{K.~O’Brien}, \bibinfo{author}{E.~Hallahan}, \bibinfo{author}{M.~A. Khan}, \bibinfo{author}{S.~Purohit}, \bibinfo{author}{U.~S. Prashanth}, \bibinfo{author}{E.~Raff}, et~al.,
\newblock \bibinfo{title}{Pythia: A suite for analyzing large language models across training and scaling},
\newblock in: \bibinfo{booktitle}{International Conference on Machine Learning}, \bibinfo{organization}{PMLR}, pp. \bibinfo{pages}{2397--2430}.
%Type = Misc
\bibitem[{Conover et~al.(2023)Conover, Hayes, Mathur, Xie, Wan, Shah, Ghodsi, Wendell, Zaharia, and Xin}]{DatabricksBlog2023DollyV2}
\bibinfo{author}{M.~Conover}, \bibinfo{author}{M.~Hayes}, \bibinfo{author}{A.~Mathur}, \bibinfo{author}{J.~Xie}, \bibinfo{author}{J.~Wan}, \bibinfo{author}{S.~Shah}, \bibinfo{author}{A.~Ghodsi}, \bibinfo{author}{P.~Wendell}, \bibinfo{author}{M.~Zaharia}, \bibinfo{author}{R.~Xin}, \bibinfo{title}{Free dolly: Introducing the world's first truly open instruction-tuned llm}, \bibinfo{year}{2023}.
%Type = Misc
\bibitem[{Taori et~al.(2023)Taori, Gulrajani, Zhang, Dubois, Li, Guestrin, Liang, and Hashimoto}]{alpaca}
\bibinfo{author}{R.~Taori}, \bibinfo{author}{I.~Gulrajani}, \bibinfo{author}{T.~Zhang}, \bibinfo{author}{Y.~Dubois}, \bibinfo{author}{X.~Li}, \bibinfo{author}{C.~Guestrin}, \bibinfo{author}{P.~Liang}, \bibinfo{author}{T.~B. Hashimoto}, \bibinfo{title}{Stanford alpaca: An instruction-following llama model}, \bibinfo{howpublished}{\url{https://github.com/tatsu-lab/stanford_alpaca}}, \bibinfo{year}{2023}.
%Type = Misc
\bibitem[{Chiang et~al.(2023)Chiang, Li, Lin, Sheng, Wu, Zhang, Zheng, Zhuang, Zhuang, Gonzalez, Stoica, and Xing}]{vicuna2023}
\bibinfo{author}{W.-L. Chiang}, \bibinfo{author}{Z.~Li}, \bibinfo{author}{Z.~Lin}, \bibinfo{author}{Y.~Sheng}, \bibinfo{author}{Z.~Wu}, \bibinfo{author}{H.~Zhang}, \bibinfo{author}{L.~Zheng}, \bibinfo{author}{S.~Zhuang}, \bibinfo{author}{Y.~Zhuang}, \bibinfo{author}{J.~E. Gonzalez}, \bibinfo{author}{I.~Stoica}, \bibinfo{author}{E.~P. Xing}, \bibinfo{title}{Vicuna: An open-source chatbot impressing gpt-4 with 90\%* chatgpt quality}, \bibinfo{year}{2023}.
%Type = Misc
\bibitem[{Team(2024)}]{dbrx}
\bibinfo{author}{T.~M.~R. Team}, \bibinfo{title}{Introducing dbrx: A new state-of-the-art open llm}, \bibinfo{year}{2024}.
%Type = Article
\bibitem[{Rich(1985)}]{rich1985artificial}
\bibinfo{author}{E.~Rich},
\newblock \bibinfo{title}{Artificial intelligence and the humanities},
\newblock \bibinfo{journal}{Computers and the Humanities} \bibinfo{volume}{19} (\bibinfo{year}{1985}) \bibinfo{pages}{117--122}.
%Type = Article
\bibitem[{Venkatasubramanian(1989)}]{venkatasubramanian1989cache}
\bibinfo{author}{V.~Venkatasubramanian},
\newblock \bibinfo{title}{Knowledge-based systems in process engineering: Case studies in heuristic classification},
\newblock \bibinfo{journal}{Austin, TX: The CACHE Corporation}  (\bibinfo{year}{1989}).
%Type = Article
\bibitem[{Ungar and Venkatasubramanian(1990)}]{venkatasubramanian1990cache}
\bibinfo{author}{L.~Ungar}, \bibinfo{author}{V.~Venkatasubramanian},
\newblock \bibinfo{title}{Advance reasoning architectures for expert systems},
\newblock \bibinfo{journal}{Austin, TX: The CACHE Corporation}  (\bibinfo{year}{1990}).
%Type = Book
\bibitem[{Russell and Norvig(2016)}]{russell2016artificial}
\bibinfo{author}{S.~J. Russell}, \bibinfo{author}{P.~Norvig}, \bibinfo{title}{Artificial intelligence: a modern approach}, \bibinfo{publisher}{Pearson}, \bibinfo{year}{2016}.
%Type = Article
\bibitem[{Iri et~al.(1979)Iri, Aoki, O'Shima, and Matsuyama}]{iri1979algorithm}
\bibinfo{author}{M.~Iri}, \bibinfo{author}{K.~Aoki}, \bibinfo{author}{E.~O'Shima}, \bibinfo{author}{H.~Matsuyama},
\newblock \bibinfo{title}{An algorithm for diagnosis of system failures in the chemical process},
\newblock \bibinfo{journal}{Computers \& Chemical Engineering} \bibinfo{volume}{3} (\bibinfo{year}{1979}) \bibinfo{pages}{489--493}.
%Type = Article
\bibitem[{Vaidhyanathan and Venkatasubramanian(1995)}]{vaidhyanathan1995digraph}
\bibinfo{author}{R.~Vaidhyanathan}, \bibinfo{author}{V.~Venkatasubramanian},
\newblock \bibinfo{title}{Digraph-based models for automated hazop analysis},
\newblock \bibinfo{journal}{Reliability Engineering \& System Safety} \bibinfo{volume}{50} (\bibinfo{year}{1995}) \bibinfo{pages}{33--49}.
%Type = Article
\bibitem[{Maurya et~al.(2003)Maurya, Rengaswamy, and Venkatasubramanian}]{maurya2003systematic}
\bibinfo{author}{M.~R. Maurya}, \bibinfo{author}{R.~Rengaswamy}, \bibinfo{author}{V.~Venkatasubramanian},
\newblock \bibinfo{title}{A systematic framework for the development and analysis of signed digraphs for chemical processes. 1. algorithms and analysis},
\newblock \bibinfo{journal}{Industrial \& engineering chemistry research} \bibinfo{volume}{42} (\bibinfo{year}{2003}) \bibinfo{pages}{4789--4810}.
%Type = Article
\bibitem[{Johnsson and {\AA}rz{\'e}n(1998)}]{johnsson1998grafchart}
\bibinfo{author}{C.~Johnsson}, \bibinfo{author}{K.-E. {\AA}rz{\'e}n},
\newblock \bibinfo{title}{Grafchart and its relations to grafcet and petri nets},
\newblock \bibinfo{journal}{IFAC Proceedings Volumes} \bibinfo{volume}{31} (\bibinfo{year}{1998}) \bibinfo{pages}{95--100}.
%Type = Article
\bibitem[{Viswanathan et~al.(1998{\natexlab{a}})Viswanathan, Johnsson, Srinivasan, Venkatasubramanian, and {\"A}rzen}]{viswanathan1998automatingpart1}
\bibinfo{author}{S.~Viswanathan}, \bibinfo{author}{C.~Johnsson}, \bibinfo{author}{R.~Srinivasan}, \bibinfo{author}{V.~Venkatasubramanian}, \bibinfo{author}{K.~E. {\"A}rzen},
\newblock \bibinfo{title}{Automating operating procedure synthesis for batch processes: Part i. knowledge representation and planning framework},
\newblock \bibinfo{journal}{Computers \& chemical engineering} \bibinfo{volume}{22} (\bibinfo{year}{1998}{\natexlab{a}}) \bibinfo{pages}{1673--1685}.
%Type = Article
\bibitem[{Viswanathan et~al.(1998{\natexlab{b}})Viswanathan, Johnsson, Srinivasan, Venkatasubramanian, and {\"A}rzen}]{viswanathan1998automatingpart2}
\bibinfo{author}{S.~Viswanathan}, \bibinfo{author}{C.~Johnsson}, \bibinfo{author}{R.~Srinivasan}, \bibinfo{author}{V.~Venkatasubramanian}, \bibinfo{author}{K.~E. {\"A}rzen},
\newblock \bibinfo{title}{Automating operating procedure synthesis for batch processes: Part ii. implementation and application},
\newblock \bibinfo{journal}{Computers \& chemical engineering} \bibinfo{volume}{22} (\bibinfo{year}{1998}{\natexlab{b}}) \bibinfo{pages}{1687--1698}.
%Type = Article
\bibitem[{Banares-Alcantara et~al.(1985{\natexlab{a}})Banares-Alcantara, Sriram, Venkatasubramanian, Westerberg, and Rychener}]{banares1985knowledge}
\bibinfo{author}{R.~Banares-Alcantara}, \bibinfo{author}{D.~Sriram}, \bibinfo{author}{V.~Venkatasubramanian}, \bibinfo{author}{A.~Westerberg}, \bibinfo{author}{M.~Rychener},
\newblock \bibinfo{title}{Knowledge-based expert systems for cad},
\newblock \bibinfo{journal}{Chemical engineering progress} \bibinfo{volume}{81} (\bibinfo{year}{1985}{\natexlab{a}}) \bibinfo{pages}{25--30}.
%Type = Article
\bibitem[{Banares-Alcantara et~al.(1985{\natexlab{b}})Banares-Alcantara, Westerberg, and Rychener}]{banares1985development}
\bibinfo{author}{R.~Banares-Alcantara}, \bibinfo{author}{A.~W. Westerberg}, \bibinfo{author}{M.~D. Rychener},
\newblock \bibinfo{title}{Development of an expert system for physical property predictions},
\newblock \bibinfo{journal}{Computers \& chemical engineering} \bibinfo{volume}{9} (\bibinfo{year}{1985}{\natexlab{b}}) \bibinfo{pages}{127--142}.
%Type = Article
\bibitem[{Banares-Alcantara et~al.(1987)Banares-Alcantara, Westerberg, Ko, and Rychener}]{banares1987decade}
\bibinfo{author}{R.~Banares-Alcantara}, \bibinfo{author}{A.~Westerberg}, \bibinfo{author}{E.~Ko}, \bibinfo{author}{M.~Rychener},
\newblock \bibinfo{title}{Decade—a hybrid expert system for catalyst selection—i. expert system consideration},
\newblock \bibinfo{journal}{Computers \& chemical engineering} \bibinfo{volume}{11} (\bibinfo{year}{1987}) \bibinfo{pages}{265--277}.
%Type = Article
\bibitem[{Banares-Alcantara et~al.(1988)Banares-Alcantara, Ko, Westerberg, and Rychener}]{banares1988decade}
\bibinfo{author}{R.~Banares-Alcantara}, \bibinfo{author}{E.~Ko}, \bibinfo{author}{A.~Westerberg}, \bibinfo{author}{M.~Rychener},
\newblock \bibinfo{title}{Decade—a hybrid expert system for catalyst selection—ii. final architecture and results},
\newblock \bibinfo{journal}{Computers \& Chemical Engineering} \bibinfo{volume}{12} (\bibinfo{year}{1988}) \bibinfo{pages}{923--938}.
%Type = Article
\bibitem[{Rich and Venkatasubramanian(1987)}]{rich1987model}
\bibinfo{author}{S.~H. Rich}, \bibinfo{author}{V.~Venkatasubramanian},
\newblock \bibinfo{title}{Model-based reasoning in diagnostic expert systems for chemical process plants},
\newblock \bibinfo{journal}{Computers \& Chemical Engineering} \bibinfo{volume}{11} (\bibinfo{year}{1987}) \bibinfo{pages}{111--122}.
%Type = Article
\bibitem[{Venkatasubramanian and Rich(1988)}]{venkatasubramanian1988object}
\bibinfo{author}{V.~Venkatasubramanian}, \bibinfo{author}{S.~Rich},
\newblock \bibinfo{title}{An object-oriented two-tier architecture for integrating compiled and deep-level knowledge for process diagnosis},
\newblock \bibinfo{journal}{Computers \& Chemical Engineering} \bibinfo{volume}{12} (\bibinfo{year}{1988}) \bibinfo{pages}{903--921}.
%Type = Inproceedings
\bibitem[{Aldea et~al.(2003)Aldea, Ba{\~n}ares-Alc{\'a}ntara, Bocio, Gramajo, Isern, Kokossis, Jim{\'e}nez, Moreno, and Ria{\~n}o}]{aldea2003ontology}
\bibinfo{author}{A.~Aldea}, \bibinfo{author}{R.~Ba{\~n}ares-Alc{\'a}ntara}, \bibinfo{author}{J.~Bocio}, \bibinfo{author}{J.~Gramajo}, \bibinfo{author}{D.~Isern}, \bibinfo{author}{A.~C. Kokossis}, \bibinfo{author}{L.~Jim{\'e}nez}, \bibinfo{author}{A.~Moreno}, \bibinfo{author}{D.~Ria{\~n}o},
\newblock \bibinfo{title}{An ontology-based knowledge management platform.},
\newblock in: \bibinfo{booktitle}{IIWeb}, pp. \bibinfo{pages}{177--182}.
%Type = Article
\bibitem[{Venkatasubramanian et~al.(2006)Venkatasubramanian, Zhao, Joglekar, Jain, Hailemariam, Suresh, Akkisetty, Morris, and Reklaitis}]{venkatasubramanian2006ontological}
\bibinfo{author}{V.~Venkatasubramanian}, \bibinfo{author}{C.~Zhao}, \bibinfo{author}{G.~Joglekar}, \bibinfo{author}{A.~Jain}, \bibinfo{author}{L.~Hailemariam}, \bibinfo{author}{P.~Suresh}, \bibinfo{author}{P.~Akkisetty}, \bibinfo{author}{K.~Morris}, \bibinfo{author}{G.~V. Reklaitis},
\newblock \bibinfo{title}{Ontological informatics infrastructure for pharmaceutical product development and manufacturing},
\newblock \bibinfo{journal}{Computers \& chemical engineering} \bibinfo{volume}{30} (\bibinfo{year}{2006}) \bibinfo{pages}{1482--1496}.
%Type = Book
\bibitem[{Marquardt et~al.(2010)Marquardt, Morbach, Wiesner, Yang, and CAPE}]{marquardt2010re}
\bibinfo{author}{W.~Marquardt}, \bibinfo{author}{J.~Morbach}, \bibinfo{author}{A.~Wiesner}, \bibinfo{author}{A.~Yang}, \bibinfo{author}{O.~CAPE}, \bibinfo{title}{A Re-Usable Ontology for Chemical Process Engineering}, \bibinfo{publisher}{Springer}, \bibinfo{year}{2010}.
%Type = Article
\bibitem[{Hailemariam and Venkatasubramanian(2010{\natexlab{a}})}]{hailemariam2010purdue1}
\bibinfo{author}{L.~Hailemariam}, \bibinfo{author}{V.~Venkatasubramanian},
\newblock \bibinfo{title}{Purdue ontology for pharmaceutical engineering: part i. conceptual framework},
\newblock \bibinfo{journal}{Journal of Pharmaceutical Innovation} \bibinfo{volume}{5} (\bibinfo{year}{2010}{\natexlab{a}}) \bibinfo{pages}{88--99}.
%Type = Article
\bibitem[{Hailemariam and Venkatasubramanian(2010{\natexlab{b}})}]{hailemariam2010purdue2}
\bibinfo{author}{L.~Hailemariam}, \bibinfo{author}{V.~Venkatasubramanian},
\newblock \bibinfo{title}{Purdue ontology for pharmaceutical engineering: Part ii. applications},
\newblock \bibinfo{journal}{Journal of Pharmaceutical Innovation} \bibinfo{volume}{5} (\bibinfo{year}{2010}{\natexlab{b}}) \bibinfo{pages}{139--146}.
%Type = Article
\bibitem[{Katare and Venkatasubramanian(2001)}]{katare2001agent}
\bibinfo{author}{S.~Katare}, \bibinfo{author}{V.~Venkatasubramanian},
\newblock \bibinfo{title}{An agent-based learning framework for modeling microbial growth},
\newblock \bibinfo{journal}{Engineering applications of artificial intelligence} \bibinfo{volume}{14} (\bibinfo{year}{2001}) \bibinfo{pages}{715--726}.
%Type = Article
\bibitem[{Julka et~al.(2002)Julka, Srinivasan, and Karimi}]{julka2002agent}
\bibinfo{author}{N.~Julka}, \bibinfo{author}{R.~Srinivasan}, \bibinfo{author}{I.~Karimi},
\newblock \bibinfo{title}{Agent-based supply chain management—1: framework},
\newblock \bibinfo{journal}{Computers \& Chemical Engineering} \bibinfo{volume}{26} (\bibinfo{year}{2002}) \bibinfo{pages}{1755--1769}.
%Type = Article
\bibitem[{Sundaram et~al.(2001)Sundaram, Ghosh, Caruthers, and Venkatasubramanian}]{sundaram2001design}
\bibinfo{author}{A.~Sundaram}, \bibinfo{author}{P.~Ghosh}, \bibinfo{author}{J.~M. Caruthers}, \bibinfo{author}{V.~Venkatasubramanian},
\newblock \bibinfo{title}{Design of fuel additives using neural networks and evolutionary algorithms},
\newblock \bibinfo{journal}{AIChE journal} \bibinfo{volume}{47} (\bibinfo{year}{2001}) \bibinfo{pages}{1387--1406}.
%Type = Article
\bibitem[{Viswanathan et~al.(2002)Viswanathan, Shah, and Venkatasubramanian}]{viswanathan2002hybrid}
\bibinfo{author}{S.~Viswanathan}, \bibinfo{author}{N.~Shah}, \bibinfo{author}{V.~Venkatasubramanian},
\newblock \bibinfo{title}{Hybrid framework for hazard identification and assessment in batch processes},
\newblock \bibinfo{journal}{AIChE journal} \bibinfo{volume}{48} (\bibinfo{year}{2002}) \bibinfo{pages}{1765--1774}.
%Type = Article
\bibitem[{Ghosh et~al.(2003)Ghosh, Katare, Patkar, Caruthers, Venkatasubramanian, and Walker}]{ghosh2003sulfur}
\bibinfo{author}{P.~Ghosh}, \bibinfo{author}{S.~Katare}, \bibinfo{author}{P.~Patkar}, \bibinfo{author}{J.~M. Caruthers}, \bibinfo{author}{V.~Venkatasubramanian}, \bibinfo{author}{K.~A. Walker},
\newblock \bibinfo{title}{Sulfur vulcanization of natural rubber for benzothiazole accelerated formulations: from reaction mechanisms to a rational kinetic model},
\newblock \bibinfo{journal}{Rubber chemistry and technology} \bibinfo{volume}{76} (\bibinfo{year}{2003}) \bibinfo{pages}{592--693}.
%Type = Article
\bibitem[{Rich and Venkatasubramanian(1989)}]{rich1989causality}
\bibinfo{author}{S.~H. Rich}, \bibinfo{author}{V.~Venkatasubramanian},
\newblock \bibinfo{title}{Causality-based failure-driven learning in diagnostic expert systems},
\newblock \bibinfo{journal}{AIChe journal} \bibinfo{volume}{35} (\bibinfo{year}{1989}) \bibinfo{pages}{943--950}.
%Type = Article
\bibitem[{Vedam and Venkatasubramanian(1999)}]{vedam1999pca}
\bibinfo{author}{H.~Vedam}, \bibinfo{author}{V.~Venkatasubramanian},
\newblock \bibinfo{title}{Pca-sdg based process monitoring and fault diagnosis},
\newblock \bibinfo{journal}{Control engineering practice} \bibinfo{volume}{7} (\bibinfo{year}{1999}) \bibinfo{pages}{903--917}.
%Type = Article
\bibitem[{Zhao et~al.(2005{\natexlab{a}})Zhao, Bhushan, and Venkatasubramanian}]{zhao2005phasuite1}
\bibinfo{author}{C.~Zhao}, \bibinfo{author}{M.~Bhushan}, \bibinfo{author}{V.~Venkatasubramanian},
\newblock \bibinfo{title}{Phasuite: an automated hazop analysis tool for chemical processes: part i: knowledge engineering framework},
\newblock \bibinfo{journal}{Process Safety and Environmental Protection} \bibinfo{volume}{83} (\bibinfo{year}{2005}{\natexlab{a}}) \bibinfo{pages}{509--532}.
%Type = Article
\bibitem[{Zhao et~al.(2005{\natexlab{b}})Zhao, Bhushan, and Venkatasubramanian}]{zhao2005phasuite2}
\bibinfo{author}{C.~Zhao}, \bibinfo{author}{M.~Bhushan}, \bibinfo{author}{V.~Venkatasubramanian},
\newblock \bibinfo{title}{Phasuite: An automated hazop analysis tool for chemical processes: Part ii: Implementation and case study},
\newblock \bibinfo{journal}{Process Safety and Environmental Protection} \bibinfo{volume}{83} (\bibinfo{year}{2005}{\natexlab{b}}) \bibinfo{pages}{533--548}.
%Type = Article
\bibitem[{Caruthers et~al.(2003)Caruthers, Lauterbach, Thomson, Venkatasubramanian, Snively, Bhan, Katare, and Oskarsdottir}]{caruthers2003catalyst}
\bibinfo{author}{J.~Caruthers}, \bibinfo{author}{J.~A. Lauterbach}, \bibinfo{author}{K.~Thomson}, \bibinfo{author}{V.~Venkatasubramanian}, \bibinfo{author}{C.~Snively}, \bibinfo{author}{A.~Bhan}, \bibinfo{author}{S.~Katare}, \bibinfo{author}{G.~Oskarsdottir},
\newblock \bibinfo{title}{Catalyst design: knowledge extraction from high-throughput experimentation},
\newblock \bibinfo{journal}{Journal of Catalysis} \bibinfo{volume}{216} (\bibinfo{year}{2003}) \bibinfo{pages}{98--109}.
%Type = Article
\bibitem[{Katare et~al.(2004)Katare, Caruthers, Delgass, and Venkatasubramanian}]{katare2004intelligent}
\bibinfo{author}{S.~Katare}, \bibinfo{author}{J.~M. Caruthers}, \bibinfo{author}{W.~N. Delgass}, \bibinfo{author}{V.~Venkatasubramanian},
\newblock \bibinfo{title}{An intelligent system for reaction kinetic modeling and catalyst design},
\newblock \bibinfo{journal}{Industrial \& engineering chemistry research} \bibinfo{volume}{43} (\bibinfo{year}{2004}) \bibinfo{pages}{3484--3512}.
%Type = Article
\bibitem[{Hsu et~al.(2008)Hsu, Krishnamurthy, Rao, Zhao, Jagannathan, and Venkatasubramanian}]{hsu2008domain}
\bibinfo{author}{S.-H. Hsu}, \bibinfo{author}{B.~Krishnamurthy}, \bibinfo{author}{P.~Rao}, \bibinfo{author}{C.~Zhao}, \bibinfo{author}{S.~Jagannathan}, \bibinfo{author}{V.~Venkatasubramanian},
\newblock \bibinfo{title}{A domain-specific compiler theory based framework for automated reaction network generation},
\newblock \bibinfo{journal}{Computers \& Chemical Engineering} \bibinfo{volume}{32} (\bibinfo{year}{2008}) \bibinfo{pages}{2455--2470}.
%Type = Article
\bibitem[{Suresh et~al.(2010{\natexlab{a}})Suresh, Hsu, Akkisetty, Reklaitis, and Venkatasubramanian}]{suresh2010ontomodel1}
\bibinfo{author}{P.~Suresh}, \bibinfo{author}{S.-H. Hsu}, \bibinfo{author}{P.~Akkisetty}, \bibinfo{author}{G.~V. Reklaitis}, \bibinfo{author}{V.~Venkatasubramanian},
\newblock \bibinfo{title}{Ontomodel: ontological mathematical modeling knowledge management in pharmaceutical product development, 1: conceptual framework},
\newblock \bibinfo{journal}{Industrial \& Engineering Chemistry Research} \bibinfo{volume}{49} (\bibinfo{year}{2010}{\natexlab{a}}) \bibinfo{pages}{7758--7767}.
%Type = Article
\bibitem[{Suresh et~al.(2010{\natexlab{b}})Suresh, Hsu, Reklaitis, and Venkatasubramanian}]{suresh2010ontomodel2}
\bibinfo{author}{P.~Suresh}, \bibinfo{author}{S.-H. Hsu}, \bibinfo{author}{G.~V. Reklaitis}, \bibinfo{author}{V.~Venkatasubramanian},
\newblock \bibinfo{title}{Ontomodel: ontological mathematical modeling knowledge management in pharmaceutical product development, 2: applications},
\newblock \bibinfo{journal}{Industrial \& Engineering Chemistry Research} \bibinfo{volume}{49} (\bibinfo{year}{2010}{\natexlab{b}}) \bibinfo{pages}{7768--7781}.
%Type = Article
\bibitem[{Decardi-Nelson et~al.(2024)Decardi-Nelson, Alshehri, Ajagekar, and You}]{decardi2024generative}
\bibinfo{author}{B.~Decardi-Nelson}, \bibinfo{author}{A.~S. Alshehri}, \bibinfo{author}{A.~Ajagekar}, \bibinfo{author}{F.~You},
\newblock \bibinfo{title}{Generative ai and process systems engineering: The next frontier},
\newblock \bibinfo{journal}{arXiv preprint arXiv:2402.10977}  (\bibinfo{year}{2024}).
%Type = Article
\bibitem[{Hu et~al.(2021)Hu, Shen, Wallis, Allen-Zhu, Li, Wang, Wang, and Chen}]{hu2021lora}
\bibinfo{author}{E.~J. Hu}, \bibinfo{author}{Y.~Shen}, \bibinfo{author}{P.~Wallis}, \bibinfo{author}{Z.~Allen-Zhu}, \bibinfo{author}{Y.~Li}, \bibinfo{author}{S.~Wang}, \bibinfo{author}{L.~Wang}, \bibinfo{author}{W.~Chen},
\newblock \bibinfo{title}{Lora: Low-rank adaptation of large language models},
\newblock \bibinfo{journal}{arXiv preprint arXiv:2106.09685}  (\bibinfo{year}{2021}).
%Type = Article
\bibitem[{Wei et~al.(2023)Wei, Wei, Tay, Tran, Webson, Lu, Chen, Liu, Huang, Zhou et~al.}]{wei2023larger}
\bibinfo{author}{J.~Wei}, \bibinfo{author}{J.~Wei}, \bibinfo{author}{Y.~Tay}, \bibinfo{author}{D.~Tran}, \bibinfo{author}{A.~Webson}, \bibinfo{author}{Y.~Lu}, \bibinfo{author}{X.~Chen}, \bibinfo{author}{H.~Liu}, \bibinfo{author}{D.~Huang}, \bibinfo{author}{D.~Zhou}, et~al.,
\newblock \bibinfo{title}{Larger language models do in-context learning differently},
\newblock \bibinfo{journal}{arXiv preprint arXiv:2303.03846}  (\bibinfo{year}{2023}).
%Type = Article
\bibitem[{Lewis et~al.(2020)Lewis, Perez, Piktus, Petroni, Karpukhin, Goyal, K{\"u}ttler, Lewis, Yih, Rockt{\"a}schel et~al.}]{lewis2020retrieval}
\bibinfo{author}{P.~Lewis}, \bibinfo{author}{E.~Perez}, \bibinfo{author}{A.~Piktus}, \bibinfo{author}{F.~Petroni}, \bibinfo{author}{V.~Karpukhin}, \bibinfo{author}{N.~Goyal}, \bibinfo{author}{H.~K{\"u}ttler}, \bibinfo{author}{M.~Lewis}, \bibinfo{author}{W.-t. Yih}, \bibinfo{author}{T.~Rockt{\"a}schel}, et~al.,
\newblock \bibinfo{title}{Retrieval-augmented generation for knowledge-intensive nlp tasks},
\newblock \bibinfo{journal}{Advances in Neural Information Processing Systems} \bibinfo{volume}{33} (\bibinfo{year}{2020}) \bibinfo{pages}{9459--9474}.
%Type = Article
\bibitem[{Von~Stosch et~al.(2014)Von~Stosch, Oliveira, Peres, and de~Azevedo}]{von2014hybrid}
\bibinfo{author}{M.~Von~Stosch}, \bibinfo{author}{R.~Oliveira}, \bibinfo{author}{J.~Peres}, \bibinfo{author}{S.~F. de~Azevedo},
\newblock \bibinfo{title}{Hybrid semi-parametric modeling in process systems engineering: Past, present and future},
\newblock \bibinfo{journal}{Computers \& Chemical Engineering} \bibinfo{volume}{60} (\bibinfo{year}{2014}) \bibinfo{pages}{86--101}.
%Type = Inproceedings
\bibitem[{Ribeiro et~al.(2016)Ribeiro, Singh, and Guestrin}]{ribeiro2016should}
\bibinfo{author}{M.~T. Ribeiro}, \bibinfo{author}{S.~Singh}, \bibinfo{author}{C.~Guestrin},
\newblock \bibinfo{title}{" why should i trust you?" explaining the predictions of any classifier},
\newblock in: \bibinfo{booktitle}{Proceedings of the 22nd ACM SIGKDD international conference on knowledge discovery and data mining}, pp. \bibinfo{pages}{1135--1144}.
%Type = Article
\bibitem[{Lundberg and Lee(2017)}]{lundberg2017unified}
\bibinfo{author}{S.~M. Lundberg}, \bibinfo{author}{S.-I. Lee},
\newblock \bibinfo{title}{A unified approach to interpreting model predictions},
\newblock \bibinfo{journal}{Advances in neural information processing systems} \bibinfo{volume}{30} (\bibinfo{year}{2017}).
%Type = Article
\bibitem[{Wilson and Sahinidis(2017)}]{wilson2017alamo}
\bibinfo{author}{Z.~T. Wilson}, \bibinfo{author}{N.~V. Sahinidis},
\newblock \bibinfo{title}{The alamo approach to machine learning},
\newblock \bibinfo{journal}{Computers \& Chemical Engineering} \bibinfo{volume}{106} (\bibinfo{year}{2017}) \bibinfo{pages}{785--795}.
%Type = Article
\bibitem[{Udrescu and Tegmark(2020)}]{udrescu2020ai}
\bibinfo{author}{S.-M. Udrescu}, \bibinfo{author}{M.~Tegmark},
\newblock \bibinfo{title}{Ai feynman: A physics-inspired method for symbolic regression},
\newblock \bibinfo{journal}{Science Advances} \bibinfo{volume}{6} (\bibinfo{year}{2020}) \bibinfo{pages}{eaay2631}.
%Type = Incollection
\bibitem[{Jul-Rasmussen et~al.(2023)Jul-Rasmussen, Chakraborty, Venkatasubramanian, Liang, and Huusom}]{jul2023identifying}
\bibinfo{author}{P.~Jul-Rasmussen}, \bibinfo{author}{A.~Chakraborty}, \bibinfo{author}{V.~Venkatasubramanian}, \bibinfo{author}{X.~Liang}, \bibinfo{author}{J.~K. Huusom},
\newblock \bibinfo{title}{Identifying first-principles models for bubble column aeration using machine learning},
\newblock in: \bibinfo{booktitle}{Computer Aided Chemical Engineering}, volume~\bibinfo{volume}{52}, \bibinfo{publisher}{Elsevier}, \bibinfo{year}{2023}, pp. \bibinfo{pages}{1089--1094}.
%Type = Article
\bibitem[{Jul-Rasmussen et~al.(2024)Jul-Rasmussen, Chakraborty, Venkatasubramanian, Liang, and Huusom}]{jul2024hybrid}
\bibinfo{author}{P.~Jul-Rasmussen}, \bibinfo{author}{A.~Chakraborty}, \bibinfo{author}{V.~Venkatasubramanian}, \bibinfo{author}{X.~Liang}, \bibinfo{author}{J.~K. Huusom},
\newblock \bibinfo{title}{Hybrid ai modeling techniques for pilot scale bubble column aeration: A comparative study},
\newblock \bibinfo{journal}{Computers \& Chemical Engineering}  (\bibinfo{year}{2024}) \bibinfo{pages}{108655}.
%Type = Article
\bibitem[{Weininger(1988)}]{weininger1988smiles}
\bibinfo{author}{D.~Weininger},
\newblock \bibinfo{title}{Smiles, a chemical language and information system. 1. introduction to methodology and encoding rules},
\newblock \bibinfo{journal}{Journal of chemical information and computer sciences} \bibinfo{volume}{28} (\bibinfo{year}{1988}) \bibinfo{pages}{31--36}.
%Type = Article
\bibitem[{Weininger et~al.(1989)Weininger, Weininger, and Weininger}]{weininger1989smiles}
\bibinfo{author}{D.~Weininger}, \bibinfo{author}{A.~Weininger}, \bibinfo{author}{J.~L. Weininger},
\newblock \bibinfo{title}{Smiles. 2. algorithm for generation of unique smiles notation},
\newblock \bibinfo{journal}{Journal of chemical information and computer sciences} \bibinfo{volume}{29} (\bibinfo{year}{1989}) \bibinfo{pages}{97--101}.
%Type = Article
\bibitem[{Lipman and Pearson(1985)}]{lipman1985rapid}
\bibinfo{author}{D.~J. Lipman}, \bibinfo{author}{W.~R. Pearson},
\newblock \bibinfo{title}{Rapid and sensitive protein similarity searches},
\newblock \bibinfo{journal}{Science} \bibinfo{volume}{227} (\bibinfo{year}{1985}) \bibinfo{pages}{1435--1441}.
%Type = Article
\bibitem[{Pearson and Lipman(1988)}]{pearson1988improved}
\bibinfo{author}{W.~R. Pearson}, \bibinfo{author}{D.~J. Lipman},
\newblock \bibinfo{title}{Improved tools for biological sequence comparison.},
\newblock \bibinfo{journal}{Proceedings of the National Academy of Sciences} \bibinfo{volume}{85} (\bibinfo{year}{1988}) \bibinfo{pages}{2444--2448}.
%Type = Article
\bibitem[{Guo et~al.(2023)Guo, Guo, Liang, Guo, Chawla, Wiest, Zhang et~al.}]{guo2023indeed}
\bibinfo{author}{T.~Guo}, \bibinfo{author}{K.~Guo}, \bibinfo{author}{Z.~Liang}, \bibinfo{author}{Z.~Guo}, \bibinfo{author}{N.~V. Chawla}, \bibinfo{author}{O.~Wiest}, \bibinfo{author}{X.~Zhang}, et~al.,
\newblock \bibinfo{title}{What indeed can gpt models do in chemistry? a comprehensive benchmark on eight tasks},
\newblock \bibinfo{journal}{arXiv preprint arXiv:2305.18365}  (\bibinfo{year}{2023}).
%Type = Article
\bibitem[{Bran et~al.(2023)Bran, Cox, White, and Schwaller}]{bran2023chemcrow}
\bibinfo{author}{A.~M. Bran}, \bibinfo{author}{S.~Cox}, \bibinfo{author}{A.~D. White}, \bibinfo{author}{P.~Schwaller},
\newblock \bibinfo{title}{Chemcrow: Augmenting large-language models with chemistry tools},
\newblock \bibinfo{journal}{arXiv preprint arXiv:2304.05376}  (\bibinfo{year}{2023}).
%Type = Article
\bibitem[{Shi et~al.(2023)Shi, Zhang, Zhang, Liu, and Wang}]{shi2023relm}
\bibinfo{author}{Y.~Shi}, \bibinfo{author}{A.~Zhang}, \bibinfo{author}{E.~Zhang}, \bibinfo{author}{Z.~Liu}, \bibinfo{author}{X.~Wang},
\newblock \bibinfo{title}{Relm: Leveraging language models for enhanced chemical reaction prediction},
\newblock \bibinfo{journal}{arXiv preprint arXiv:2310.13590}  (\bibinfo{year}{2023}).
%Type = Article
\bibitem[{Ito et~al.(2024)Ito, Muraoka, and Nakayama}]{ito2024novo}
\bibinfo{author}{S.~Ito}, \bibinfo{author}{K.~Muraoka}, \bibinfo{author}{A.~Nakayama},
\newblock \bibinfo{title}{De novo design of organic structure-directing agents for zeolites using a general-purpose large language model},
\newblock \bibinfo{journal}{ChemRxiv}  (\bibinfo{year}{2024}).
%Type = Article
\bibitem[{Achiam et~al.(2023)Achiam, Adler, Agarwal, Ahmad, Akkaya, Aleman, Almeida, Altenschmidt, Altman, Anadkat et~al.}]{achiam2023gpt}
\bibinfo{author}{J.~Achiam}, \bibinfo{author}{S.~Adler}, \bibinfo{author}{S.~Agarwal}, \bibinfo{author}{L.~Ahmad}, \bibinfo{author}{I.~Akkaya}, \bibinfo{author}{F.~L. Aleman}, \bibinfo{author}{D.~Almeida}, \bibinfo{author}{J.~Altenschmidt}, \bibinfo{author}{S.~Altman}, \bibinfo{author}{S.~Anadkat}, et~al.,
\newblock \bibinfo{title}{Gpt-4 technical report},
\newblock \bibinfo{journal}{arXiv preprint arXiv:2303.08774}  (\bibinfo{year}{2023}).
%Type = Article
\bibitem[{Herbertz et~al.(2015)Herbertz, Sawyer, Stauber, Gueorguieva, Driscoll, Estrem, Cleverly, Desaiah, Guba, Benhadji et~al.}]{herbertz2015clinical}
\bibinfo{author}{S.~Herbertz}, \bibinfo{author}{J.~S. Sawyer}, \bibinfo{author}{A.~J. Stauber}, \bibinfo{author}{I.~Gueorguieva}, \bibinfo{author}{K.~E. Driscoll}, \bibinfo{author}{S.~T. Estrem}, \bibinfo{author}{A.~L. Cleverly}, \bibinfo{author}{D.~Desaiah}, \bibinfo{author}{S.~C. Guba}, \bibinfo{author}{K.~A. Benhadji}, et~al.,
\newblock \bibinfo{title}{Clinical development of galunisertib (ly2157299 monohydrate), a small molecule inhibitor of transforming growth factor-beta signaling pathway},
\newblock \bibinfo{journal}{Drug design, development and therapy}  (\bibinfo{year}{2015}) \bibinfo{pages}{4479--4499}.
%Type = Article
\bibitem[{Lee et~al.(2020)Lee, Yoon, Kim, Kim, Kim, So, and Kang}]{lee2020biobert}
\bibinfo{author}{J.~Lee}, \bibinfo{author}{W.~Yoon}, \bibinfo{author}{S.~Kim}, \bibinfo{author}{D.~Kim}, \bibinfo{author}{S.~Kim}, \bibinfo{author}{C.~H. So}, \bibinfo{author}{J.~Kang},
\newblock \bibinfo{title}{Biobert: a pre-trained biomedical language representation model for biomedical text mining},
\newblock \bibinfo{journal}{Bioinformatics} \bibinfo{volume}{36} (\bibinfo{year}{2020}) \bibinfo{pages}{1234--1240}.
%Type = Inproceedings
\bibitem[{Li et~al.(2023)Li, Xu, Zhao, Guo, and Liu}]{li2023chatpathway}
\bibinfo{author}{Y.~Li}, \bibinfo{author}{H.~Xu}, \bibinfo{author}{H.~Zhao}, \bibinfo{author}{H.~Guo}, \bibinfo{author}{S.~Liu},
\newblock \bibinfo{title}{Chatpathway: Conversational large language models for biology pathway detection},
\newblock in: \bibinfo{booktitle}{NeurIPS 2023 AI for Science Workshop}.
%Type = Article
\bibitem[{Gu et~al.(2021)Gu, Tinn, Cheng, Lucas, Usuyama, Liu, Naumann, Gao, and Poon}]{gu2021domain}
\bibinfo{author}{Y.~Gu}, \bibinfo{author}{R.~Tinn}, \bibinfo{author}{H.~Cheng}, \bibinfo{author}{M.~Lucas}, \bibinfo{author}{N.~Usuyama}, \bibinfo{author}{X.~Liu}, \bibinfo{author}{T.~Naumann}, \bibinfo{author}{J.~Gao}, \bibinfo{author}{H.~Poon},
\newblock \bibinfo{title}{Domain-specific language model pretraining for biomedical natural language processing},
\newblock \bibinfo{journal}{ACM Transactions on Computing for Healthcare (HEALTH)} \bibinfo{volume}{3} (\bibinfo{year}{2021}) \bibinfo{pages}{1--23}.
%Type = Article
\bibitem[{Bagal et~al.(2021)Bagal, Aggarwal, Vinod, and Priyakumar}]{bagal2021molgpt}
\bibinfo{author}{V.~Bagal}, \bibinfo{author}{R.~Aggarwal}, \bibinfo{author}{P.~Vinod}, \bibinfo{author}{U.~D. Priyakumar},
\newblock \bibinfo{title}{Molgpt: molecular generation using a transformer-decoder model},
\newblock \bibinfo{journal}{Journal of Chemical Information and Modeling} \bibinfo{volume}{62} (\bibinfo{year}{2021}) \bibinfo{pages}{2064--2076}.
%Type = Article
\bibitem[{Beltagy et~al.(2019)Beltagy, Lo, and Cohan}]{beltagy2019scibert}
\bibinfo{author}{I.~Beltagy}, \bibinfo{author}{K.~Lo}, \bibinfo{author}{A.~Cohan},
\newblock \bibinfo{title}{Scibert: A pretrained language model for scientific text},
\newblock \bibinfo{journal}{arXiv preprint arXiv:1903.10676}  (\bibinfo{year}{2019}).
%Type = Article
\bibitem[{Zhang et~al.(2024)Zhang, Ding, Lyv, Wang, Yin, Zhang, Yu, Wang, Li, Xiang et~al.}]{zhang2024scientific}
\bibinfo{author}{Q.~Zhang}, \bibinfo{author}{K.~Ding}, \bibinfo{author}{T.~Lyv}, \bibinfo{author}{X.~Wang}, \bibinfo{author}{Q.~Yin}, \bibinfo{author}{Y.~Zhang}, \bibinfo{author}{J.~Yu}, \bibinfo{author}{Y.~Wang}, \bibinfo{author}{X.~Li}, \bibinfo{author}{Z.~Xiang}, et~al.,
\newblock \bibinfo{title}{Scientific large language models: A survey on biological \& chemical domains},
\newblock \bibinfo{journal}{arXiv preprint arXiv:2401.14656}  (\bibinfo{year}{2024}).
%Type = Inproceedings
\bibitem[{Venkatasubramanian(2019)}]{venkatasubramanian2019artificial}
\bibinfo{author}{V.~Venkatasubramanian},
\newblock \bibinfo{title}{Artificial intelligence in materials science: the good, the bad, and the ugly},
\newblock in: \bibinfo{booktitle}{2019 AIChE Annual Meeting}, \bibinfo{organization}{AIChE}.
%Type = Misc
\bibitem[{AIChE(2019)}]{vvyoutubechiang2019}
\bibinfo{author}{AIChE}, \bibinfo{title}{Venkat venkatasubramanian on artificial intelligence in chemical engineering}, \bibinfo{year}{2019}.
%Type = Article
\bibitem[{Psichogios and Ungar(1992)}]{psichogios1992hybrid}
\bibinfo{author}{D.~C. Psichogios}, \bibinfo{author}{L.~H. Ungar},
\newblock \bibinfo{title}{A hybrid neural network-first principles approach to process modeling},
\newblock \bibinfo{journal}{AIChE Journal} \bibinfo{volume}{38} (\bibinfo{year}{1992}) \bibinfo{pages}{1499--1511}.
%Type = Article
\bibitem[{Trinh et~al.(2024)Trinh, Wu, Le, He, and Luong}]{trinh2024solving}
\bibinfo{author}{T.~H. Trinh}, \bibinfo{author}{Y.~Wu}, \bibinfo{author}{Q.~V. Le}, \bibinfo{author}{H.~He}, \bibinfo{author}{T.~Luong},
\newblock \bibinfo{title}{Solving olympiad geometry without human demonstrations},
\newblock \bibinfo{journal}{Nature} \bibinfo{volume}{625} (\bibinfo{year}{2024}) \bibinfo{pages}{476--482}.
%Type = Article
\bibitem[{Roose(2024)}]{Roose2024NYT}
\bibinfo{author}{K.~Roose},
\newblock \bibinfo{title}{A.i.’s black boxes just got a little less mysterious},
\newblock \bibinfo{journal}{The New York Times}  (\bibinfo{year}{2024}).
%Type = Article
\bibitem[{Templeton et~al.(2024)Templeton, Conerly, Marcus, Lindsey, Bricken, Chen, Pearce, Citro, Ameisen, Jones, Cunningham, Turner, McDougall, MacDiarmid, Freeman, Sumers, Rees, Batson, Jermyn, Carter, Olah, and Henighan}]{templeton2024scaling}
\bibinfo{author}{A.~Templeton}, \bibinfo{author}{T.~Conerly}, \bibinfo{author}{J.~Marcus}, \bibinfo{author}{J.~Lindsey}, \bibinfo{author}{T.~Bricken}, \bibinfo{author}{B.~Chen}, \bibinfo{author}{A.~Pearce}, \bibinfo{author}{C.~Citro}, \bibinfo{author}{E.~Ameisen}, \bibinfo{author}{A.~Jones}, \bibinfo{author}{H.~Cunningham}, \bibinfo{author}{N.~L. Turner}, \bibinfo{author}{C.~McDougall}, \bibinfo{author}{M.~MacDiarmid}, \bibinfo{author}{C.~D. Freeman}, \bibinfo{author}{T.~R. Sumers}, \bibinfo{author}{E.~Rees}, \bibinfo{author}{J.~Batson}, \bibinfo{author}{A.~Jermyn}, \bibinfo{author}{S.~Carter}, \bibinfo{author}{C.~Olah}, \bibinfo{author}{T.~Henighan},
\newblock \bibinfo{title}{Scaling monosemanticity: Extracting interpretable features from claude 3 sonnet},
\newblock \bibinfo{journal}{Transformer Circuits Thread}  (\bibinfo{year}{2024}).
%Type = Article
\bibitem[{Roose(2023)}]{Roose2023NYT}
\bibinfo{author}{K.~Roose},
\newblock \bibinfo{title}{A conversation with bing’s chatbot left me deeply unsettled},
\newblock \bibinfo{journal}{The New York Times}  (\bibinfo{year}{2023}).
%Type = Misc
\bibitem[{Gunasekar et~al.(2023)Gunasekar, Zhang, Aneja, Cesar, Mendes, Giorno, Gopi, Javaheripi, Kauffmann, de~Rosa, Saarikivi, Salim, Shah, Singh~Behl, Wang, Bubeck, Eldan, Kalai, Lee, and Li}]{gunasekar2023textbooks}
\bibinfo{author}{S.~Gunasekar}, \bibinfo{author}{Y.~Zhang}, \bibinfo{author}{J.~Aneja}, \bibinfo{author}{C.~Cesar}, \bibinfo{author}{T.~Mendes}, \bibinfo{author}{A.~D. Giorno}, \bibinfo{author}{S.~Gopi}, \bibinfo{author}{M.~Javaheripi}, \bibinfo{author}{P.~Kauffmann}, \bibinfo{author}{G.~de~Rosa}, \bibinfo{author}{O.~Saarikivi}, \bibinfo{author}{A.~Salim}, \bibinfo{author}{S.~Shah}, \bibinfo{author}{H.~Singh~Behl}, \bibinfo{author}{X.~Wang}, \bibinfo{author}{S.~Bubeck}, \bibinfo{author}{R.~Eldan}, \bibinfo{author}{A.~T. Kalai}, \bibinfo{author}{Y.~T. Lee}, \bibinfo{author}{Y.~Li}, \bibinfo{title}{Textbooks are all you need}, \bibinfo{year}{2023}.
%Type = Article
\bibitem[{Abdin et~al.(2024)Abdin, Jacobs, Awan, Aneja, Awadallah, Awadalla, Bach, Bahree, Bakhtiari, Behl et~al.}]{abdin2024phi}
\bibinfo{author}{M.~Abdin}, \bibinfo{author}{S.~A. Jacobs}, \bibinfo{author}{A.~A. Awan}, \bibinfo{author}{J.~Aneja}, \bibinfo{author}{A.~Awadallah}, \bibinfo{author}{H.~Awadalla}, \bibinfo{author}{N.~Bach}, \bibinfo{author}{A.~Bahree}, \bibinfo{author}{A.~Bakhtiari}, \bibinfo{author}{H.~Behl}, et~al.,
\newblock \bibinfo{title}{Phi-3 technical report: A highly capable language model locally on your phone},
\newblock \bibinfo{journal}{arXiv preprint arXiv:2404.14219}  (\bibinfo{year}{2024}).
%Type = Article
\bibitem[{Ji et~al.(2023)Ji, Lee, Frieske, Yu, Su, Xu, Ishii, Bang, Madotto, and Fung}]{ji2023survey}
\bibinfo{author}{Z.~Ji}, \bibinfo{author}{N.~Lee}, \bibinfo{author}{R.~Frieske}, \bibinfo{author}{T.~Yu}, \bibinfo{author}{D.~Su}, \bibinfo{author}{Y.~Xu}, \bibinfo{author}{E.~Ishii}, \bibinfo{author}{Y.~J. Bang}, \bibinfo{author}{A.~Madotto}, \bibinfo{author}{P.~Fung},
\newblock \bibinfo{title}{Survey of hallucination in natural language generation},
\newblock \bibinfo{journal}{ACM Computing Surveys} \bibinfo{volume}{55} (\bibinfo{year}{2023}) \bibinfo{pages}{1--38}.
%Type = Article
\bibitem[{Petroni et~al.(2019)Petroni, Rockt{\"a}schel, Lewis, Bakhtin, Wu, Miller, and Riedel}]{petroni2019language}
\bibinfo{author}{F.~Petroni}, \bibinfo{author}{T.~Rockt{\"a}schel}, \bibinfo{author}{P.~Lewis}, \bibinfo{author}{A.~Bakhtin}, \bibinfo{author}{Y.~Wu}, \bibinfo{author}{A.~H. Miller}, \bibinfo{author}{S.~Riedel},
\newblock \bibinfo{title}{Language models as knowledge bases?},
\newblock \bibinfo{journal}{arXiv preprint arXiv:1909.01066}  (\bibinfo{year}{2019}).
%Type = Article
\bibitem[{Zhang et~al.(2022)Zhang, Song, Li, Zhou, and Song}]{zhang2022survey}
\bibinfo{author}{H.~Zhang}, \bibinfo{author}{H.~Song}, \bibinfo{author}{S.~Li}, \bibinfo{author}{M.~Zhou}, \bibinfo{author}{D.~Song},
\newblock \bibinfo{title}{A survey of controllable text generation using transformer-based pre-trained language models},
\newblock \bibinfo{journal}{arXiv preprint arXiv:2201.05337}  (\bibinfo{year}{2022}).
%Type = Article
\bibitem[{Pan et~al.(2023)Pan, Luo, Wang, Chen, Wang, and Wu}]{pan2023unifying}
\bibinfo{author}{S.~Pan}, \bibinfo{author}{L.~Luo}, \bibinfo{author}{Y.~Wang}, \bibinfo{author}{C.~Chen}, \bibinfo{author}{J.~Wang}, \bibinfo{author}{X.~Wu},
\newblock \bibinfo{title}{Unifying large language models and knowledge graphs: A roadmap},
\newblock \bibinfo{journal}{arXiv preprint arXiv:2306.08302}  (\bibinfo{year}{2023}).

\end{thebibliography}

\end{document}